\documentclass{article}

\PassOptionsToPackage{numbers, compress}{natbib}
\usepackage{hyperref}       %

    \usepackage[final]{neurips_2020}

\usepackage[utf8]{inputenc} %
\usepackage[T1]{fontenc}    %
\usepackage{url}            %
\usepackage{booktabs}       %
\usepackage{amsfonts}       %
\usepackage{amsmath}
\usepackage{amssymb}
\usepackage{amsthm}
\usepackage{amscd}
\usepackage{mathtools}
\usepackage{bm}
\usepackage{algorithmic}
\usepackage{algorithm}
\usepackage{wrapfig}
\mathtoolsset{showonlyrefs=true}
\usepackage{nicefrac}       %
\usepackage{microtype}      %
\usepackage[dvipsnames]{xcolor}
\usepackage{arydshln}
\usepackage{placeins}

\newtheorem{theorem}{Theorem}

\newtheorem{cor}{Corollary}

\title{Munchausen Reinforcement Learning}

\author{
 Nino Vieillard \\
 Google Research, Brain Team \\
 Universit\'e de Lorraine, CNRS, Inria, IECL\\
 F-54000 Nancy, France\\
 \texttt{vieillard@google.com}
 \And
 Olivier Pietquin \\
 Google Research, Brain Team \\
 \texttt{pietquin@google.com}
 \And 
 Matthieu Geist \\
 Google Research, Brain Team \\
 \texttt{mfgeist@google.com}}

\newcommand{\gr}{\mathcal{G}}
\newcommand{\E}{\mathbb{E}}
\newcommand{\h}{\mathcal{H}}

\DeclareMathOperator*{\argmax}{argmax}
\DeclareMathOperator*{\softmax}{sm}

\DeclareMathOperator{\kl}{KL}
\DeclareMathOperator{\gapop}{\delta}

\newcommand{\states}{\mathcal{S}}
\newcommand{\actions}{\mathcal{A}}

\newcommand{\abs}[1]{\left\lvert#1\right\rvert}
\newcommand{\gap}{\gapop}
\DeclareMathOperator{\FC}{FC}
\DeclareMathOperator{\Conv}{Conv}
\newcommand{\un}{\mathbf{1}}

\begin{document}

\maketitle

\begin{abstract}
    Bootstrapping is a core mechanism in Reinforcement Learning (RL). Most algorithms, based on temporal differences, replace the true value of a transiting state by their current estimate of this value. Yet, another estimate could be leveraged to bootstrap RL: the current policy. Our core contribution stands in a very simple idea: adding the scaled log-policy to the immediate reward. We show that slightly modifying Deep $Q$-Network (DQN) in that way provides an agent that is competitive with distributional methods on Atari games, without making use of distributional RL, $n$-step returns or prioritized replay. To demonstrate the versatility of this idea, we also use it together with an Implicit Quantile Network (IQN). The resulting agent outperforms Rainbow on Atari, installing a new State of the Art with very little modifications to the original algorithm. To add to this empirical study, we provide strong theoretical insights on what happens under the hood -- implicit Kullback-Leibler regularization and increase of the action-gap.
\end{abstract}

\section{Introduction}

 Most Reinforcement Learning (RL) algorithms make use of Temporal Difference (TD) learning~\citep{sutton1988learning} in some ways. It is a well-known bootstrapping mechanism that consists in replacing the unknown true value of a transiting state by its current estimate and use it as a target for learning. Yet, agents compute another estimate while learning that could be leveraged to bootstrap RL: their current policy. Indeed, it reflects the agent's hunch about which actions should be executed next and thus, which actions are good. Building upon this observation, our core contribution stands in a very simple idea: optimizing for the \emph{immediate} reward  \emph{augmented} by the scaled log-policy of the agent when using any TD scheme. We insist right away that this is different from maximum entropy RL~\citep{ziebart2010modeling}, that \emph{subtracts} the scaled log-policy to \emph{all} rewards, and aims at maximizing both the expected return and the expected entropy of the resulting policy. We call this general approach ``\emph{Munchausen} Reinforcement Learning'' (M-RL), as a reference to a famous passage of \textit{The Surprising Adventures of Baron Munchausen} by~\citet{raspe:1785}, where the Baron pulls himself out of a swamp by pulling on his own hair.

To demonstrate the genericity and the strength of this idea, we introduce it into the most popular RL agent: the seminal Deep $Q$-Network (DQN)~\citep{mnih2015human}. Yet, DQN does not compute stochastic policies, which prevents using log-policies. So, we first introduce a straightforward generalization of DQN to maximum entropy RL~\cite{ziebart2010modeling,haarnoja2018soft}, and then modify the resulting TD update by adding the scaled log-policy to the immediate reward. The resulting algorithm, referred to as Munchausen-DQN (M-DQN), is thus genuinely a slight modification of DQN. Yet, it comes with  strong empirical performances. On the Arcade Learning Environment (ALE)~\cite{bellemare2013arcade}, not only it surpasses the original DQN by a large margin, but it also overtakes C51~\cite{bellemare2017distributional}, the first agent based on distributional RL (distRL). As far as we know, M-DQN is the first agent not using distRL that outperforms a distRL agent\footnote{It appears that the benefits of distRL do not really come from RL principles, but rather from the regularizing effect of modelling a distribution and its role as an auxiliary task in a deep learning
context~\cite{lyle2019comparative}.}. The current state of the art for single agent algorithms is considered to be Rainbow~\citep{hessel2018rainbow}, that combines C51 with other enhacements to DQN, and does not rely on massivly distributed computation (unlike R2D2~\cite{kapturowski2018recurrent}, SEED~\cite{Espeholt2020SEED} or Agent57~\cite{badia2020agent57}). To demonstrate the versatility of the M-RL idea, we apply the same recipe to modify Implicit Quantile Network (IQN)~\cite{dabney2018implicit}, a recent distRL agent. The resulting Munchausen-IQN (M-IQN) surpasses Rainbow, installing a new state of the art.

To support these empirical results, we provide strong theoretical insights about what happens under the hood. We rewrite M-DQN under an abstract dynamic programming scheme and show that it implicitly performs Kullback-Leibler (KL) regularization between consecutive policies. M-RL is not the first approach to take advantage of KL regularization~\cite{schulman2015trust,abdolmaleki2018maximum}, but we show that, because this regularization is implicit, it comes with stronger theoretical guarantees. From this, we link M-RL to Conservative Value Iteration (CVI)~\cite{kozuno2019theoretical} and Dynamic Policy Programming (DPP)~\cite{azar2012dynamic} that were not introduced with deep RL implementations.
 We also draw connections with Advantage Learning (AL)~\cite{baird1999reinforcement,bellemare2016increasing} and study the effect of M-RL on the action-gap~\cite{farahmand2011action}. While M-RL is not the first scheme to induce an increase of the action-gap~\cite{bellemare2016increasing}, it is the first one that allows quantifying this increase.

\section{Munchausen Reinforcement Learning \label{sec:mrl}}

RL is usually formalized within the Markov Decision Processes (MDP) framework. An MDP models the environment and is a tuple $\{\states, \actions, P, r, \gamma\}$, with $\states$ and $\actions$ the state and action spaces, $P$ the Markovian transition kernel, $r$ the bounded reward function and $\gamma$ the discount factor.  The RL agent interacts with the MDP using a policy $\pi$, that associates to every state either an action (deterministic policy) or a distribution over actions (stochastic policy). The quality of this interaction is quantified by the expected discounted cumulative return, formalized as the state-action value function, $q_\pi(s,a) = \mathbb{E}_\pi[\sum_{t=0}^\infty \gamma^t r(s_t, a_t) | s_0=s, a_0 = a]$, the expectation being over trajectories induced by the policy $\pi$ and the dynamics $P$. An optimal policy satisfies $\pi_*\in\argmax_{\pi} q_\pi$. The associated optimal value function $q_* = q_{\pi_*}$ satisfies the Bellman equation $q_*(s,a) = r(s,a) + \gamma E_{s'|s,a}[\max_{a'}q_*(s',a')]$. A deterministic \textit{greedy} policy satisfies $\pi(a|s)=1$ for $a\in\argmax_{a'}q(s,a')$
and will be written $\pi\in\gr(q)$. We also use \textit{softmax} policies, $\pi=\softmax(q) \Leftrightarrow \pi(a|s) = \frac{\exp q(s,a)}{\sum_{a'} \exp q(s,a')}$.

A standard RL agent maintains both a $q$-function and a policy (that can be implicit, for example $\pi\in\gr(q)$), and it aims at learning an optimal policy. To do so, it often relies on Temporal Difference (TD) updates. To recall the principle of TD learning, we quickly revisit the classical $Q$-learning algorithm~\cite{watkins1992q}. When interacting with the environment the agent observes transitions $(s_t,a_t,r_t,s_{t+1})$. Would the optimal $q$-function $q_*$ be known in the state $s_{t+1}$, the agent could use it as a learning target and build successive estimates as $q(s_t,a_t) \leftarrow q(s_t,a_t) + \eta (r_t + \gamma \max_{a'} q_*(s_{t+1},a') - q(s_t,a_t))$, using the Bellman equation, $\eta$ being a learning rate. Yet, $q_*$ is unknown, and the agent actually uses its current estimate $q$ instead, which is known as bootstrapping. 

We argue that the $q$-function is not the sole quantity that could be used to bootstrap RL. Let's assume that an optimal deterministic policy $\pi_*$ is known. The log-policy is therefore $0$ for optimal actions, and $-\infty$ for sub-optimal ones. This is a very strong learning signal, that we could add to the reward to ease learning, without changing the optimal control. The optimal policy $\pi_*$ being obviously unknown, we replace it by the agent's current estimate $\pi$, and we assume stochastic policies for numerical stability. To sum up, M-RL is a very simple idea, that consists in replacing $r_t$ by $r_t+\alpha \ln \pi(a_t|s_t)$ in any TD scheme, assuming that the current agent's policy $\pi$ is stochastic, so as to bootstrap the current agent's guess about what actions are good.

To demonstrate the generality of this approach, we use it to enhance the seminal DQN~\cite{mnih2015human} deep RL algorithm. In DQN, the $q$-values are estimated by an online $Q$-network $q_\theta$, with weights copied regularly to a target network $q_{\bar\theta}$. The agent behaves following a policy $\pi_\theta \in \gr(q_\theta)$ (with $\varepsilon$-greedy exploration), and stores transitions $(s_t,a_t,r_t, s_{t+1})$ in a FIFO replay buffer $\mathcal{B}$. DQN performs stochastic gradient descent on the loss $\hat{\mathbb{E}}_\mathcal{B}[(q_\theta(s_t,a_t) - \hat{q}_\text{dqn}(r_t, s_{t+1}))^2]$, regressing the target $\hat{q}_\text{dqn}$:
\begin{equation}
\label{eq:dqn_target}
    \hat{q}_\text{dqn}(r_t, s_{t+1}) = r_t + \gamma \sum_{a'\in\actions} \pi_{\bar{\theta}}(a'|s_{t+1}) q_{\bar\theta}(s_{t+1}, a') \text{ with } \pi_{\bar{\theta}}\in\gr(q_{\bar{\theta}}).
\end{equation}
To derive Munchausen-DQN (M-DQN), we simply modify the regression target. M-RL assumes stochastic policies while DQN computes  deterministic policies. A simple way to address this is to not only maximize the return, but also the entropy of the resulting policy, that is adopting the viewpoint of maximum entropy RL~\cite{ziebart2010modeling,haarnoja2018soft}. It is straightforward to extend DQN to this setting, see Appx.~\ref{subappx:Soft-DQN} for a detailed derivation. We call the resulting agent Soft-DQN (S-DQN). Let $\tau$ be the temperature parameter scaling the entropy, it just amounts to replace the original regression target by
\begin{equation}
\label{eq:soft_dqn_target}
    \hat{q}_\text{s-dqn}(r_t, s_{t+1}) = r_t + \gamma \sum_{a' \in \actions} \pi_{\bar{\theta}}(a'|s_{t+1}) \Big(q_{\bar\theta}(s_{t+1}, a') \textcolor{blue}{- \tau \ln \pi_{\bar\theta}(a'|s_{t+1})}\Big) \text{ with } \textcolor{blue}{\pi_{\bar{\theta}}=\softmax(\frac{q_{\bar\theta}}{\tau}}),
\end{equation}
where we highlighted the differences with DQN in \textcolor{blue}{blue}. Notice that this is nothing more than the most straightforward discrete-actions version of Soft Actor-Critic (SAC)~\cite{haarnoja2018soft}. Notice also that in the limit $\tau\rightarrow 0$ we retrieve DQN. The last step to obtain M-DQN is to add the scaled log-policy to the reward.
Let $\alpha\in[0,1]$ be a scaling factor, the regression target of M-DQN is thus
\begin{equation}
    \hat{q}_\text{m-dqn}(r_t, s_{t+1}) = r_t \textcolor{red}{+\alpha\tau\ln\pi_{\bar{\theta}}(a_t|s_t)} + \gamma \sum_{a' \in \actions} \pi_{\bar{\theta}}(a'|s_{t+1}) \Big(q_{\bar\theta}(s_{t+1}, a') \textcolor{blue}{- \tau \ln \pi_{\bar\theta}(a'|s_{t+1})}\Big) ,
    \label{eq:munchausen_dqn_target}
\end{equation}
still with $\textcolor{blue}{\pi_{\bar{\theta}}}=\textcolor{blue}{\softmax(\frac{q_{\bar\theta}}{\tau})}$,
where we highlighted the difference with Soft-DQN in \textcolor{red}{red} (retrieved by setting $\alpha=0$). Hence, M-DQN is genuinely obtained by replacing $\hat{q}_\text{dqn}$ by 
$\hat{q}_\text{m-dqn}$ as the regression target of DQN. All details of the resulting algorithm are provide in Appx.~\ref{subappx:m-agent}.

\begin{figure}[thb]
    \centering
    \includegraphics[width=.49\linewidth]{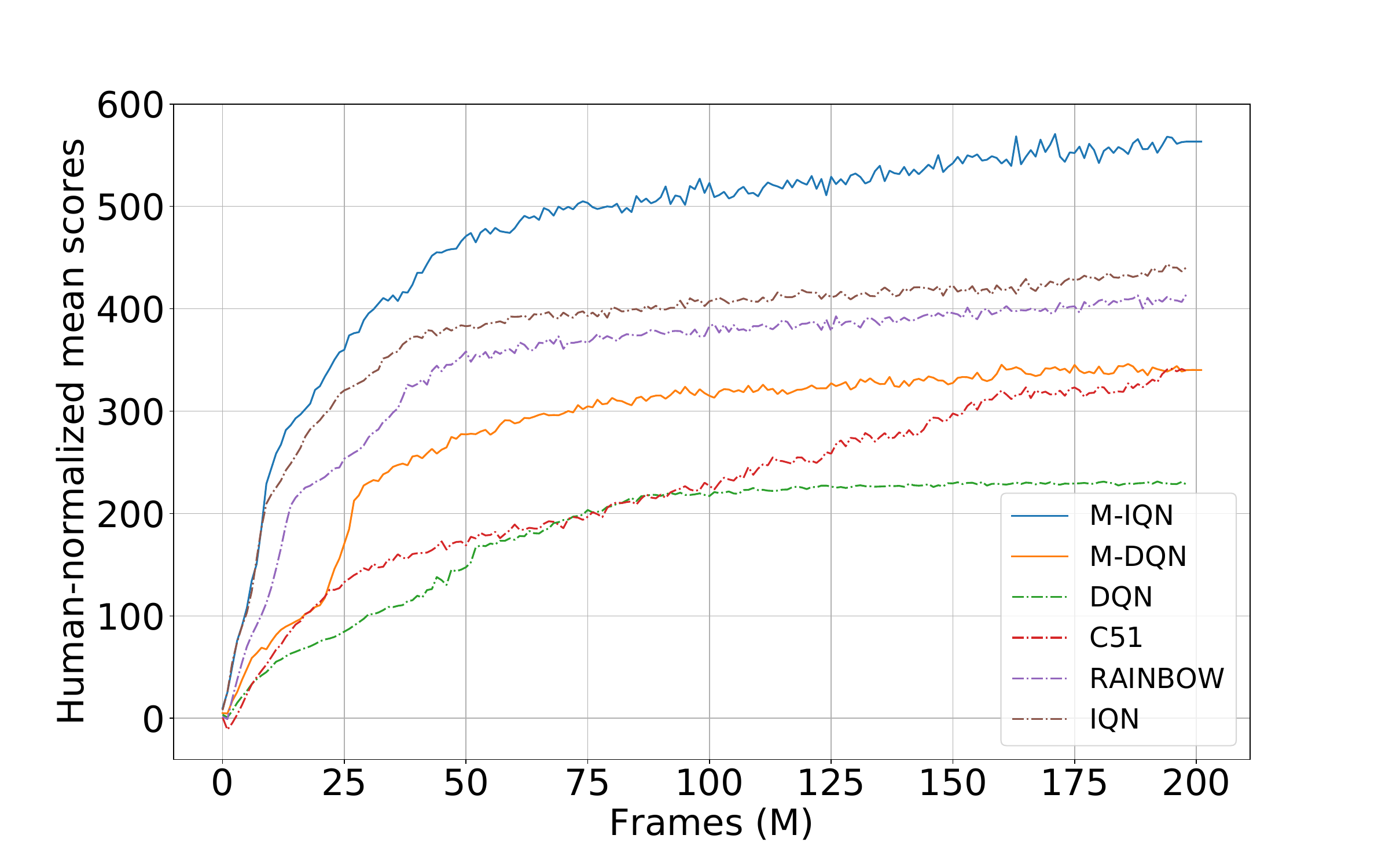}
    \includegraphics[width=.49\linewidth]{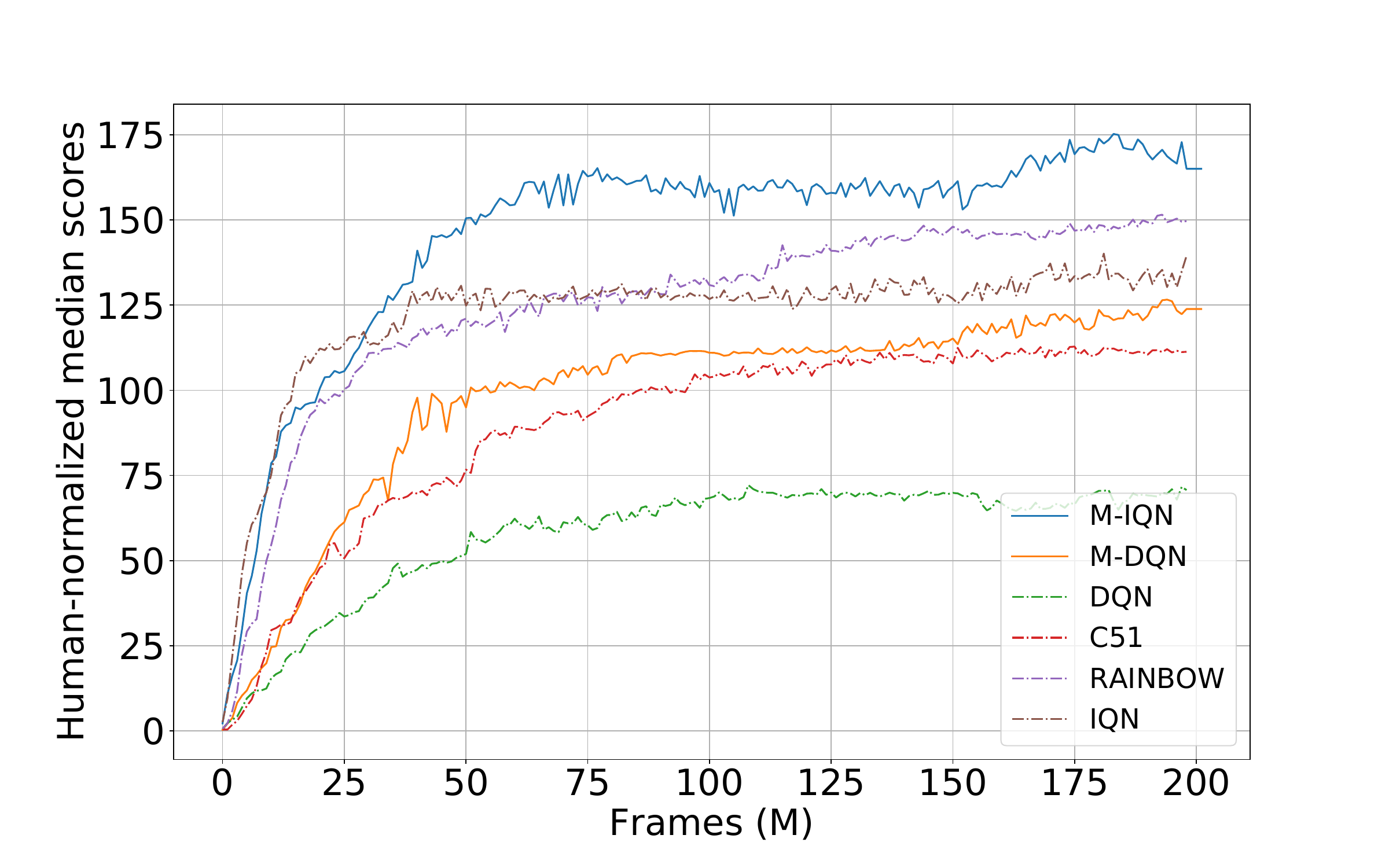}
    \caption{\textbf{Left:} Human-normalized mean scores. \textbf{Right:} Human-normalized median scores.
    }
    \label{fig:human_normalized}
\end{figure}

Despite being an extremely simple modification of DQN, M-DQN is very efficient. We show in Fig.~\ref{fig:rainbow_normalized} the Human-normalized mean and median scores for various agents on the full set of 60 Atari games of ALE (more details in Sec.~\ref{sec:experiments}). We observe that M-DQN significantly outperforms DQN, but also C51~\cite{bellemare2017distributional}. As far we know, M-DQN is the first method that is not based on distRL which overtakes C51. These are quite encouraging empirical results.

To demonstrate the versatility of the M-RL principle, we also combine it with IQN~\cite{dabney2018implicit}, a recent and efficient distRL agent (note that IQN has had recent successors, such as Fully Parameterized Quantile Function (FQF)~\cite{yang2019fully}, to which in principle, we could also apply M-RL). We denote the resulting algorithm M-IQN. In a nutshell, IQN does not estimate the $q$-function, but the distribution of which the $q$-function is the mean, using a distributional Bellman operator. The (implicit) policy is still greedy according to the $q$-function, computed as the (empirical) mean of the estimated distribution. We apply the exact same recipe: derive soft-IQN using the principle of maximum entropy RL (which is as easy as for DQN), and add the scaled log-policy to the reward. For the sake of showing the generality of our method, we combine M-RL with a version of IQN that uses $3$-steps returns (and we compare to IQN and Rainbow, that both use the same). We can observe on Fig.~\ref{fig:rainbow_normalized} that M-IQN outperforms Rainbow, both in terms of mean and median scores, and thus defines the new state of the art. In addition, even when using only $1$-step returns, M-IQN still outperforms Rainbow. This result and the details of M-IQN can be found respectively in Appx.~\ref{subappx:n_steps} and~\ref{subappx:m-agent}.

\section{What happens under the hood? \label{sec:what}}

The impressive empirical results of M-RL (see Sec.~\ref{sec:experiments} for more) call for some theoretical insights. To provide them, we frame M-DQN in an abstract Approximate Dynamic Programming (ADP) framework and analyze it.
We mainly provide two strong results: (1) M-DQN implicitly performs KL regularization between successive policies, which translates in an averaging effect of approximation errors (instead of accumulation in general ADP frameworks); (2) it increases the action-gap by a quantifiable amount which also helps dealing with approximation errors.
We also use this section to draw connections with the existing literature in ADP. Let's first introduce some additional notations.

We write $\Delta_X$ the simplex over the finite set $X$ and $Y^X$ the set of applications from $X$ to the set $Y$. With this, an MDP is $\{\states, \actions, P\in\Delta_{\states}^{\states\times\actions}, r\in\mathbb{R}^{\states\times\actions}, \gamma\in(0,1)\}$, the state and action spaces being assumed finite. For $f,g\in\mathbb{R}^{\states\times\actions}$, we define a component-wise dot product $\langle f, g\rangle = (\sum_{a}f(s,a) g(s,a))_s\in\mathbb{R}^\states$. This will be used with $q$-functions and (log-) policies, \textit{e.g.} for expectations: $\E_{a\sim\pi(\cdot|s)}[q(s,a)] = \langle\pi,q\rangle(s)$.
For $v\in\mathbb{R}^\states$, we have $P v = (\E_{s'|s,a}[v(s')])_{s,a} = (\sum_{s'}P(s'|s,a)v(s'))_{s,a} \in\mathbb{R}^{\states\times\actions}$. We also defined a policy-induced transition kernel $P_\pi$ as $P_\pi q = P \langle \pi, q\rangle$. With these notations, the Bellman evaluation operator is $T_\pi q = r + \gamma P_\pi q$ and its unique fixed point is $q_\pi$. An optimal policy still satisfies $\pi_* \in \argmax_{\pi\in\Delta_\actions^\states} q_\pi$. The set of greedy policies can be written as $\gr(q) = \argmax_{\pi\in\Delta_\actions^\states} \langle \pi, q\rangle$. We'll also make use of the entropy of a policy, $\h(\pi) = -\langle \pi,\ln \pi\rangle$, and of the KL between two policies, $\kl(\pi_1||\pi_2) = \langle \pi_1, \ln\pi_1 - \ln\pi_2\rangle$.

A softmax is the maximizer of the Legendre-Fenchel transform of the entropy~\cite{boyd2004convex,vieillard2020leverage}, $\softmax(q) = \argmax_{\pi}\langle \pi, q\rangle + \h(\pi)$. Using this and the introduced notations, we can write M-DQN in the following abstract form (each iteration consists of a greedy step and an evaluation step):
\begin{equation}
\label{eq:munchvi}
\begin{cases}
    \pi_{k+1} = \argmax_{\pi\in\Delta_\actions^\states}\langle\pi, q_k\rangle + \textcolor{blue}{\tau\mathcal{H}(\pi)}
    \\
    q_{k+1} = r \textcolor{red}{+ \alpha\tau\ln\pi_{k+1}} + \gamma P\langle\pi_{k+1}, q_k \textcolor{blue}{- \tau \ln \pi_{k+1}}\rangle  +  \epsilon_{k+1}.
\end{cases} \quad\text{M-VI($\alpha$, $\tau$)}
\end{equation}
We call the resulting scheme Munchausen Value Iteration, or M-VI($\alpha$,$\tau$). The term $\epsilon_{k+1}$ stands for the error between the actual and the ideal update (sampling instead of expectation, approximation of $q_k$ by a neural network, fitting of the neural network). Removing the \textcolor{red}{red term}, we retrieve approximate VI (AVI) regularized by a scaled entropy, as introduced by~\citet{geist2019theory}, of which Soft-DQN is an instantiation (as well as SAC, with additional error in the greedy step). Removing also the \textcolor{blue}{blue term}, we retrieve the classic AVI~\cite{scherrer2015approximate}, of which DQN is an instantiation.

To get some insights, we rewrite the evaluation step, setting $\alpha=1$ and with $q'_k \triangleq q_k - \tau \ln \pi_k$:
\begin{align}
    q_{k+1} &= r + \tau\ln\pi_{k+1} + \gamma P\langle\pi_{k+1}, q_k - \tau \ln \pi_{k+1}\rangle  +  \epsilon_{k+1}
    \\ \Leftrightarrow
    q_{k+1} - \tau\ln\pi_{k+1} &= r + \gamma P\langle\pi_{k+1}, q_k - \tau\ln\pi_k - \tau \ln \frac{\pi_{k+1}}{\pi_k}\rangle  +  \epsilon_{k+1}
    \\ \Leftrightarrow 
    q'_{k+1} &= r + \gamma P(\langle\pi_{k+1}, q'_k \rangle - \tau \kl(\pi_{k+1}||\pi_k)) + \epsilon_{k+1}.
\end{align}
Then, the greedy step can be rewritten as (looking at what $\pi_{k+1}$ maximizes)%
\begin{align}
\label{eq:greedy-step-inline}
    \langle \pi, q_k\rangle + \tau\h(\pi)
    = \langle \pi, q'_k + \tau \ln \pi_k \rangle - \tau\langle\pi,\ln\pi\rangle
    = \langle \pi, q'_k \rangle - \tau \kl(\pi||\pi_k).
\end{align}
We have just shown that M-VI(1,$\tau$) implicitly performs KL regularization between successive policies. 

This is a very insightful result as KL regularization is the core component of recent efficient RL agents such as TRPO~\cite{schulman2015trust} or MPO~\cite{abdolmaleki2018maximum}. It is extensively discussed by \citet{vieillard2020leverage}. Interestingly, we can show that the sequence of policies produced by M-VI($\alpha$,$\tau$) is the same as the one of their Mirror Descent VI (MD-VI), with KL scaled by $\alpha\tau$ and entropy scaled by $(1-\alpha)\tau$. Thus, M-VI($\alpha$,$\tau$) is equivalent to MD-VI($\alpha\tau$, $(1-\alpha)\tau$), as formalized below (proof in Appx.~\ref{subappx:proof_equivalence}).
\begin{theorem}
\label{thm:equivalence}
    For any $k\geq 0$, define $q'_k = q_k - \alpha\tau\ln\pi_k$, we have
    \begin{equation}
        \eqref{eq:munchvi} \Leftrightarrow
        \begin{cases}
            \pi_{k+1} = \argmax_{\pi\in\Delta_\actions^\states} \langle \pi, q'_k \rangle - \alpha\tau \kl(\pi||\pi_k) + (1-\alpha)\tau \h(\pi)
            \\
            q'_{k+1} = r + \gamma P(\langle \pi_{k+1}, q'_k\rangle - \alpha\tau \kl(\pi_{k+1}||\pi_k) + (1-\alpha)\tau \h(\pi_{k+1})) + \epsilon_{k+1}
        \end{cases}.
    \end{equation}
    Moreover, \cite[Thm.~1]{vieillard2020leverage} applies to M-VI(1,$\tau$) and \cite[Thm.~2]{vieillard2020leverage} applies to M-VI($\alpha<1$,$\tau$).%
\end{theorem}

In their work, \citet{vieillard2020leverage} show that using regularization can reduce the dependency to the horizon $(1-\gamma)^{-1}$ and that using a KL divergence allows for a compensation of the errors $\epsilon_k$ over iterations, which is not true for classical ADP. We refer to them for a detailed discussion on this topic. However, we would like to highlight that they acknowledge that \emph{their theoretical analysis does not apply to the deep RL setting}. The reason being that their analysis does not hold when the greedy step is approximated, and they deem as impossible to do the greedy step exactly when using neural network.
Indeed, computing $\pi_{k+1}$ by maximizing eq.~\eqref{eq:greedy-step-inline} yields an analytical solution proportional to $\pi_k \exp(\frac{q_k}{\tau})$, and that thus depends on the previous policy $\pi_k$. Consequently, the solution to this equation cannot be computed exactly when using deep function approximation (unless one would be willing to remember every computed policy). On the contrary, 
\emph{their analysis applies in our deep RL setting}. In M-VI, the KL regularization is implicit, so \emph{we do not introduce errors in the greedy step}. To be precise, the greedy step of M-VI is only a softmax of the $q$-function, which can be computed exactly in a discrete actions setting, even when using deep networks. Their strong bounds for MD-VI therefore hold for M-VI, as formalized in Thm.~\ref{thm:equivalence}, and in particular for M-DQN.

Indeed, let $q_{\bar{\theta}_k}$ be the $k^\text{th}$ update of the target network, write $q_k = q_{\bar{\theta}_k}$, $\pi_{k+1} = \softmax(\frac{q_k}{\tau})$, and define $\epsilon_{k+1} = q_{k+1} - (r + \alpha\ln\pi_{k+1} - \gamma P \langle \pi_{k+1}, q_k - \tau \ln\pi_{k+1}\rangle)$, the difference between the actual update and the ideal one. As a direct corollary of Thm.~\ref{thm:equivalence} and~\cite[Thm.~1]{vieillard2020leverage}, we have that, for $\alpha=1$,
\begin{equation}
    \|q_* - q_{\pi_k}\|_{\infty}\leq 
    \frac{2}{1-\gamma} \left\|\frac{1}{k}\sum_{j=1}^k \epsilon_j\right\|_\infty
    + \frac{4}{(1-\gamma)^2} \frac{r_\text{max}+\tau\ln|\actions|}{k},
\end{equation}
with $r_\text{max}$ the maximum reward (in absolute value), and with $q_{\pi_k}$ the true value function of the policy of M-DQN. This is a very strong bound. The error term is $\|\frac{1}{k}\sum_{j=1}^k \epsilon_j\|_\infty$, to be compared to the one of AVI~\cite{scherrer2015approximate}, $(1-\gamma) \sum_{j=1}^k \gamma^{k-j} \|\epsilon_j\|_\infty$. Instead of having a discounted sum of the norms of the errors, we have the norm of the average of the errors. This is very interesting, as it allows for a compensation of errors between iterations instead of an accumulation (sum and norm do not commute). The error term is scaled by $(1-\gamma)^{-1}$ (the average horizon of the MDP), while the one of AVI would be scaled by $(1-\gamma)^{-2}$. This is also quite interesting, a $\gamma$ close to 1 impacts less negatively the bound. We refer to~\cite[Sec.~4.1]{vieillard2020leverage} for further discussions about the advantage of this kind of bounds. Similarly, we could derive a bound for the case $\alpha<1$, and even more general and meaningful component-wise bounds. We defer the statement of these bounds and their proofs to Appx.~\ref{subappx:bounds}.

From Eq.~\eqref{eq:munchvi}, we can also relate the proposed approach to another part of the literature. Still from basic properties of the Legendre-Fenchel transform, we have that $\max_\pi \langle q,\pi\rangle + \tau \h(\pi) = \langle\pi_{k+1},q_k\rangle + \tau\h(\pi_{k+1}) = \ln\langle 1, \exp q\rangle$. In other words, if the maximizer is the softmax, the maximum is the log-sum-$\exp$. Using this, Eq.~\eqref{eq:munchvi} can be rewritten as (see Appx.~\ref{subappx:related_works} for a detailed derivation)
\begin{equation}
    q_{k+1} = r + \gamma P(\tau \ln\langle 1, \exp\frac{q_k}{\tau}\rangle) + \alpha(q_k - \tau \ln\langle 1, \exp\frac{q_k}{\tau}\rangle) + \epsilon_{k+1}.
    \label{eq:cvi}
\end{equation}
This is very close to Conservative Value Iteration\footnote{In CVI, $\langle 1, \exp\frac{q_k}{\tau}\rangle$ is replaced by $\langle \frac{1}{|\actions|}, \exp\frac{q_k}{\tau}\rangle$.} (CVI)~\cite{kozuno2019theoretical}, a purely theoretical algorithm, as far as we know. With $\alpha=0$ (without Munchausen), we get Soft Q-learning~\cite{fox2015taming,haarnoja2017reinforcement}. Notice that with this, CVI can be seen as soft $Q$-learning plus a scaled and smooth advantage (the term $\alpha(q_k - \tau \ln\langle 1, \exp\frac{q_k}{\tau}\rangle)$). With $\alpha=1$, we retrieve a variation of Dynamic Policy Programming (DPP)~\cite[Appx.~A]{azar2012dynamic}.
DPP has been extended to a deep learning setting~\cite{tsurumine2017deep}, but it is less efficient than DQN\footnote{
In fact, \citet{tsurumine2017deep} show better performance for deep DPP than for DQN in their setting. Yet, their experiment involves a small number of interactions, while the function estimated by DPP is naturally diverging. See~\cite[Sec.~6]{vieillard2019momentum} for further discussion about this.
}~\cite{vieillard2020leverage}.
Taking the limit $\tau\rightarrow 0$, we retrieve Advantage Learning (AL)~\cite{baird1999reinforcement,bellemare2016increasing} (see Appx.~\ref{subappx:related_works}):
\begin{equation}
    q_{k+1} = r + \gamma P\langle \pi_{k+1},q_k\rangle + \alpha (q_k - \langle \pi_{k+1}, q_k\rangle) + \epsilon_{k+1} \text{ with } \pi_{k+1} \in\gr(q_k).
    \label{eq:al}
\end{equation}

AL aims at increasing the action-gap~\cite{farahmand2011action} defined as the difference, for a given state, between the (optimal) value of the optimal action and that of the suboptimal ones. The intuitive reason to want a large action-gap is that it can mitigate the undesirable effects of approximation and estimation errors made on $q$ on the induced greedy policies. \citet{bellemare2016increasing} have introduced a family of Bellman-like operators that are gap-increasing. Not only we show that M-VI is gap-increasing but we also quantify the increase. To do so, we introduce some last notations. As we explained before, with $\alpha=0$, M-VI(0, $\tau$) reduces to AVI regularized by an entropy (that is, maximum entropy RL). Without error, it is known that the resulting regularized MDP has a unique optimal policy $\pi_*^{\tau}$ and a unique optimal $q$-function\footnote{It can be related to the unregularized optimal $q$-function, $\|q_*^{\tau} - q_*\|_\infty \leq \frac{\tau\ln|\actions|}{1-\gamma}$~\cite{geist2019theory}.} $q_*^\tau$~\cite{geist2019theory}. This being defined, we can state our result (proven in Appx.~\ref{subappx:proof_action_gap}).
\begin{theorem}
\label{thm:action_gap}
    For any state $s\in\states$, define the action-gap of an MPD regularized by an entropy scaled by $\tau$ as $\gap_*^\tau(s) = \max_{a} q_*^\tau(s,a) - q_*^\tau(s,\cdot)\in\mathbb{R}_+^\actions$. Define also $\gap^{\alpha,\tau}_k(s)$ as the action-gap for the $k^\text{th}$ iteration of M-VI($\alpha$,$\tau$), without error ($\epsilon_k=0$): $\gap^{\alpha,\tau}_k(s) = \max_a q_k(s,a) - q_k(s,\cdot)\in\mathbb{R}_+^\actions$. Then, for any $s\in\states$, for any $0\leq\alpha\leq 1$ and for any $\tau>0$, we have
    \begin{equation}
        \lim_{k\rightarrow\infty} \gap^{\alpha,\tau}_k(s) = \frac{1+\alpha}{1-\alpha} \gap_*^{(1-\alpha)\tau}(s),
    \end{equation}
with the convention that $\infty\cdot 0 = 0$ for $\alpha=1$.
\end{theorem}
Thus, the original action-gap is multiplied by $\frac{1+\alpha}{1-\alpha}$ with M-VI. In the limit $\alpha=1$, it is even infinite (and zero for the optimal actions). This suggests choosing a large value of $\alpha$, but not too close to 1 (for numerical stability: if having a large action-gap is desirable, having an infinite one is not).

\section{Experiments \label{sec:experiments}}

\paragraph{Munchausen agents.}
We implement M-DQN and M-IQN as variations of  respectively DQN and IQN from Dopamine~\cite{castro2018dopamine}. We use the same hyperparameters for IQN\footnote{By default, Dopamine's IQN uses 3-steps returns. We rather consider 1-step returns, as in~\cite{dabney2018implicit}.}, and we only change the optimizer from RMSProp to Adam for DQN. This is actually not anodyne, and we study its impact in an ablation study.
We also consider a Munchausen-specific modification, \emph{log-policy clipping}. Indeed, the log-policy term is not bounded, and can cause numerical issues if the policy becomes too close to deterministic. Thus, with a hyperparameter $l_0 <0$, we replace $\tau \ln\pi(a|s)$ by $[\tau \ln\pi(a|s)]_{l_0}^0$, where $[\cdot]_x^y$ is the clipping function. For numerical stability, we use a specific log-sum-exp trick to compute the log-policy (see App.~\ref{subappx:m-agent}). Hence, we add three parameters to the modified agent: $\alpha, \tau$ and $l_0$. After some tuning on a few Atari games, we found a working zone for these parameters to be $\alpha=0.9$, $\tau=0.03$ and $l_0=-1$, used for all experiments, in M-DQN and M-IQN.
All details about the rest of the parameters can be found in  Appx.~\ref{subappx:m-agent}. DQN and IQN use $\varepsilon$-greedy policies to interact with the environment. Although M-DQN and M-IQN produce naturally stochastic policies, we use the same $\varepsilon$-greedy policies.
We discuss this further in Appx.~\ref{subappx:comparison}, where we also compare to stochastic policies.

\paragraph{Baselines.}
First, we consider both DQN and IQN, as these are the algorithms we modify. Second, we compare to C51 because, as far as we know, it has never been outperformed by a non-distRL agent before. We also consider Rainbow, as it stands for being the state-of-the-art non-distributed agent on ALE. All our baselines are taken from Dopamine. For Rainbow, this version doesn't contain all the original improvements, but only the ones deemed as the more important and efficient by~\citet{hessel2018rainbow}: $n$-steps returns and Prioritized Experience Replay (PER)~\citep{schaul2015prioritized}, on top of C51.

\paragraph{Task.} 
We evaluate our methods and the baselines in the ALE environment, \textit{i.e.} on the full set of $60$ Atari games.
Notice that it is not a ``canonical'' environment. For example, choosing to end an episode when an agent loses a life or after game-over can dramatically change the score an agent can reach (\textit{e.g.}, \cite[Fig.~4]{castro2018dopamine}). The same holds for using sticky actions, introducing stochasticity in the dynamics (\textit{e.g.}, \cite[Fig.~6]{castro2018dopamine}). Even the ROMs could be different, with unpredictable consequences (\textit{e.g.} different video encoding). Here, we follow the methodological best practices proposed by~\citet{machado2018revisiting} and instantiated in Dopamine~\cite{castro2018dopamine}, that also makes the ALE more challenging. Notably, the results we present are hardly comparable to the ones presented in the original publications of DQN~\cite{mnih2015human}, C51~\cite{bellemare2017distributional}, Rainbow~\cite{hessel2018rainbow} or IQN~\cite{dabney2018implicit}, that use a different, easier, setting. Yet, for completeness, we report results on one game (Asterix) using an ALE setting as close as possible to the original papers, in Appx.~\ref{subappx:settings}: the baseline results match the previously published ones, and M-RL still raises improvement. We also highlight that we stick to a single-agent version of the environment: we do not claim that our method can be compared to highly distributed agents, such as R2D2~\citep{kapturowski2018recurrent} or Agent57~\citep{badia2020agent57}, that use several versions of the environment in parallel, and train on a much higher number of frames (around $10$G frames vs $200$M here). Yet, we are confident that our approach could easily apply to such agents.

\paragraph{Metrics.}
All algorithms are evaluated on the same training regime (details in Appx.\ref{subappx:m-agent}), during $200$M frames, and results are averaged over $3$ seeds. As a metric for any games, we compute the ``baseline-normalized'' score, for each iteration (here, $1$M frames), normalized so that $0\%$ corresponds to a random score, and $100\%$ to the final performance of the baseline. At each iteration, the score is the undiscounted sum of rewards, averaged over the last 100 learning episodes. The normalized score is then $\frac{a - r}{|b - r|}$, with $a$ the score of the agent, $b$ the score of the baseline, and $r$ the score of a random policy. For a human baseline, the scores are those provided in Table~\ref{tab:score_all} (Appx.~\ref{subappx:metrics}), for an agent baseline the score is the one after 200M frames. With this, we provide aggregated results, showing the mean and the median over games, as learning proceeds when the baseline is the human score (\textit{e.g.}, Fig.~\ref{fig:human_normalized}), or after 200M steps with human and Rainbow baselines in Tab.~\ref{tab:score_all} (more results in Appx.~\ref{subappx:metrics}, as learning proceeds). We also compute a ``baseline-improvement'' score as $\frac{a - b}{|b - r|}$, and use it to report a per-game improvement after 200M frames (Fig.~\ref{fig:aucs}, M-Agent versus Agent, or Appx.~\ref{subappx:metrics}).

\begin{wrapfigure}{r}{0.41\textwidth}
\vspace{-25pt}
    \centering
    \includegraphics[width=.4\textwidth]{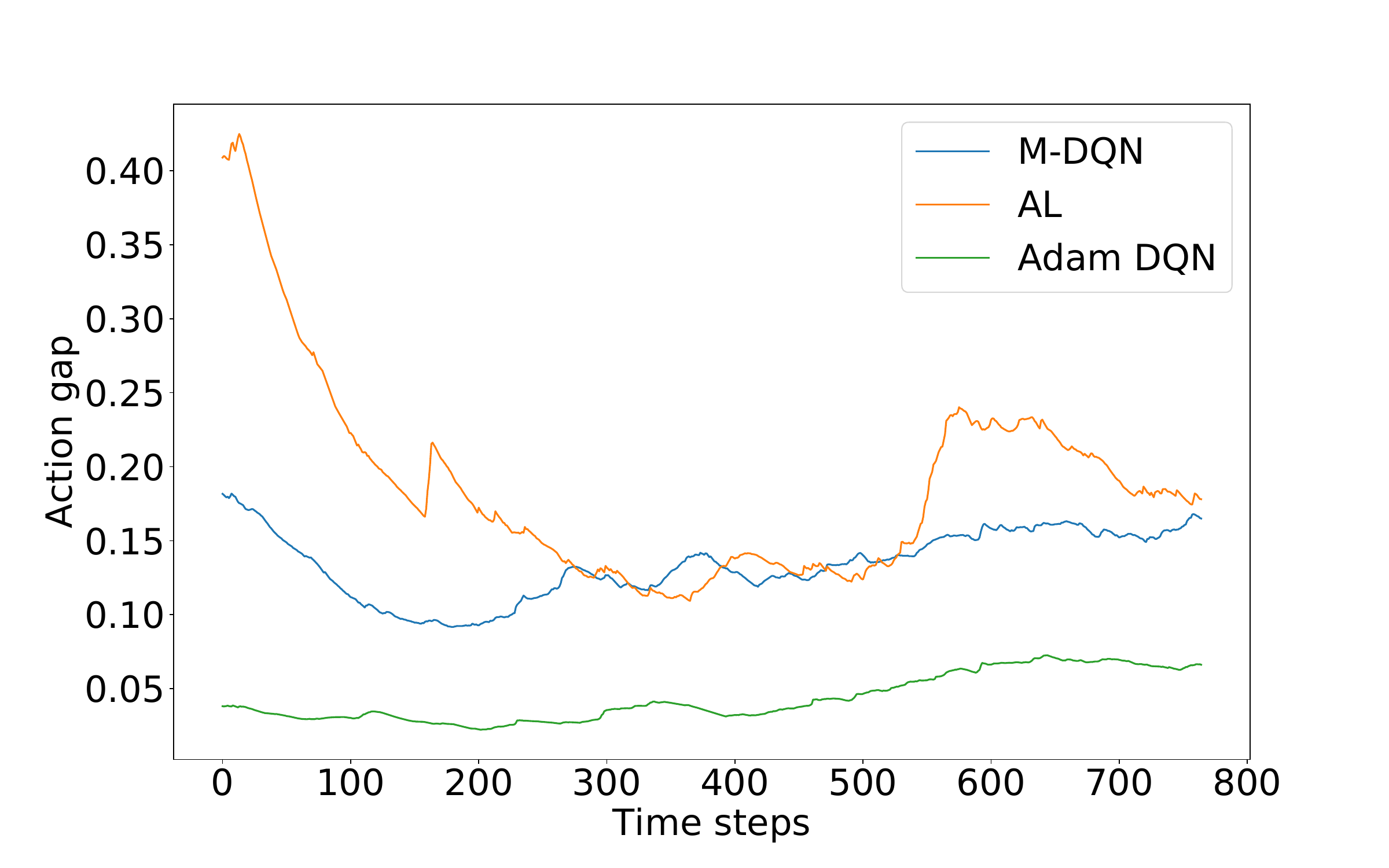}
    \caption{Action-gaps (Asterix).}
    \label{fig:action_gap}
\end{wrapfigure}

\paragraph{Action-gap.}
We start by illustrating the action-gap phenomenon suggested by Thm.~\ref{thm:action_gap}. To do so, let $q_\theta$ be the $q$-function of a given agent after training for 200M steps. At any time-step $t$, write $\hat{a}_t \in \argmax_{a\in\actions}q_\theta(s_t,a)$ the current greedy action, we compute the empirical action-gap as the difference of estimated values between the best and second best actions, $q_\theta(s_t, \hat{a}_t) - \max_{a\in\actions \setminus \{\hat{a}_t\}} q_\theta(s_t,a)$. We do so for M-DQN, for AL (that was introduced specifically to increase the action-gap) and for DQN with Adam optimizer (Adam DQN), as both build on top of it (only changing the regression targets, see Appx.~\ref{subappx:m-agent} for details). We consider the game Asterix, for which the final average performance of the agents are (roughly) 15k for Adam DQN, 13k for AL and 20k for M-DQN.
We report the results on Fig.~\ref{fig:action_gap}: we run each agent for 10 trajectories, and average the resulting action-gaps (the length of the resulting trajectory is the one of the shorter trajectory, we also apply an exponential smoothing of $0.99$).
Both M-DQN and AL increase the action-gaps compared to Adam DQN. If AL increases it more, it seems also to be less stable, and less proportional to the original action-gap. Despite this increase, it performs worse than Adam DQN (13k vs 15k), while M-DQN increases it and performs better (20k vs 15k). An explanation to this phenomenon could the one of \citet{van2019using}, who suggest that what is important is not the value of the action gap itself, but its uniformity over the state-action space: here, M-DQN seems to benefit from a more stable action-gap than AL.
This figure is for an illustrative purpose, one game is not enough to draw conclusions. Yet, the following ablation shows that globally M-DQN performs better than AL. Also, it benefits from more theoretical justifications (not only quantified action-gap increase, but also implicit KL-regularization and resulting performance bounds).

\begin{figure}
    \centering
    \includegraphics[width=0.49\textwidth]{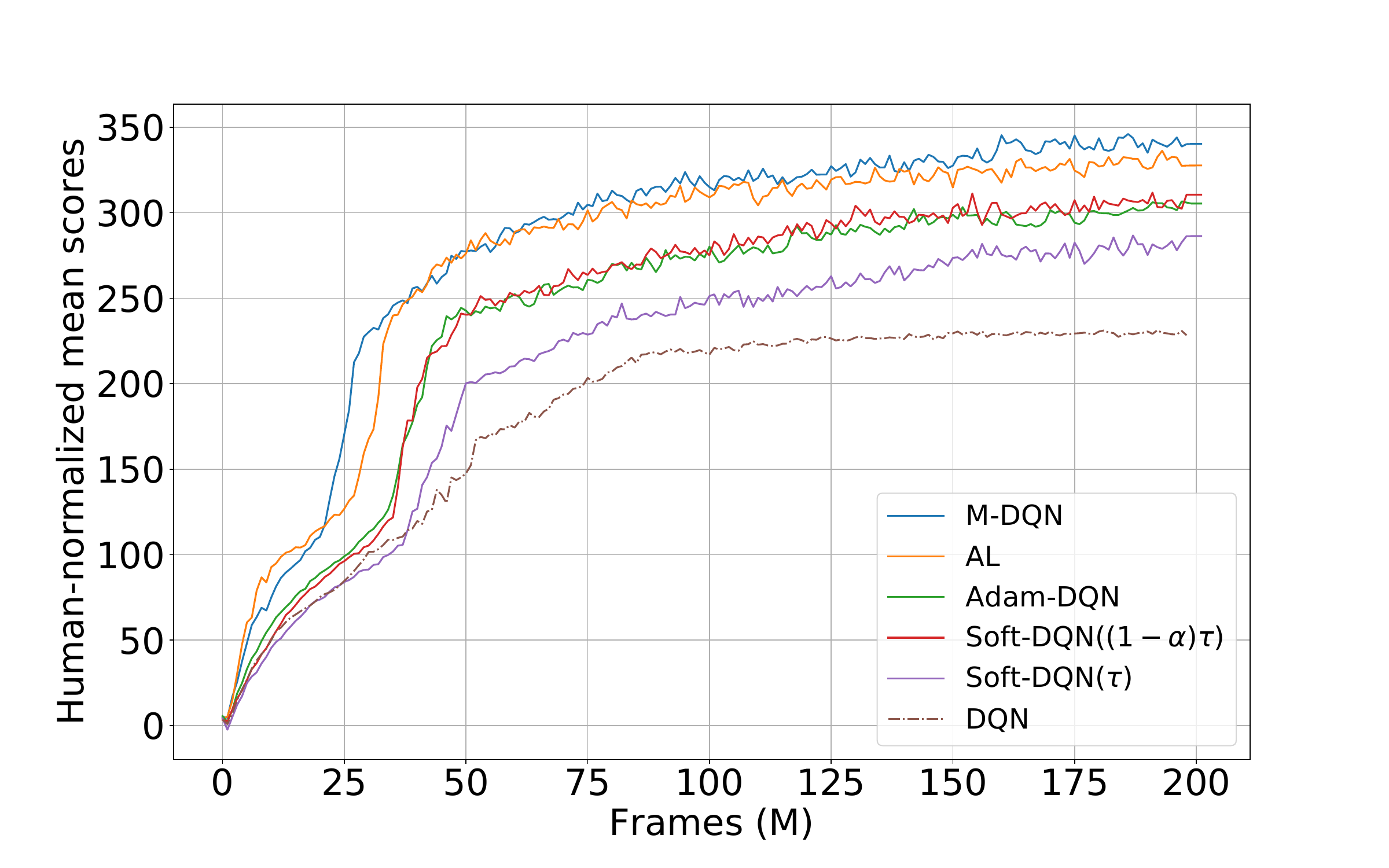}
    \includegraphics[width=0.49\textwidth]{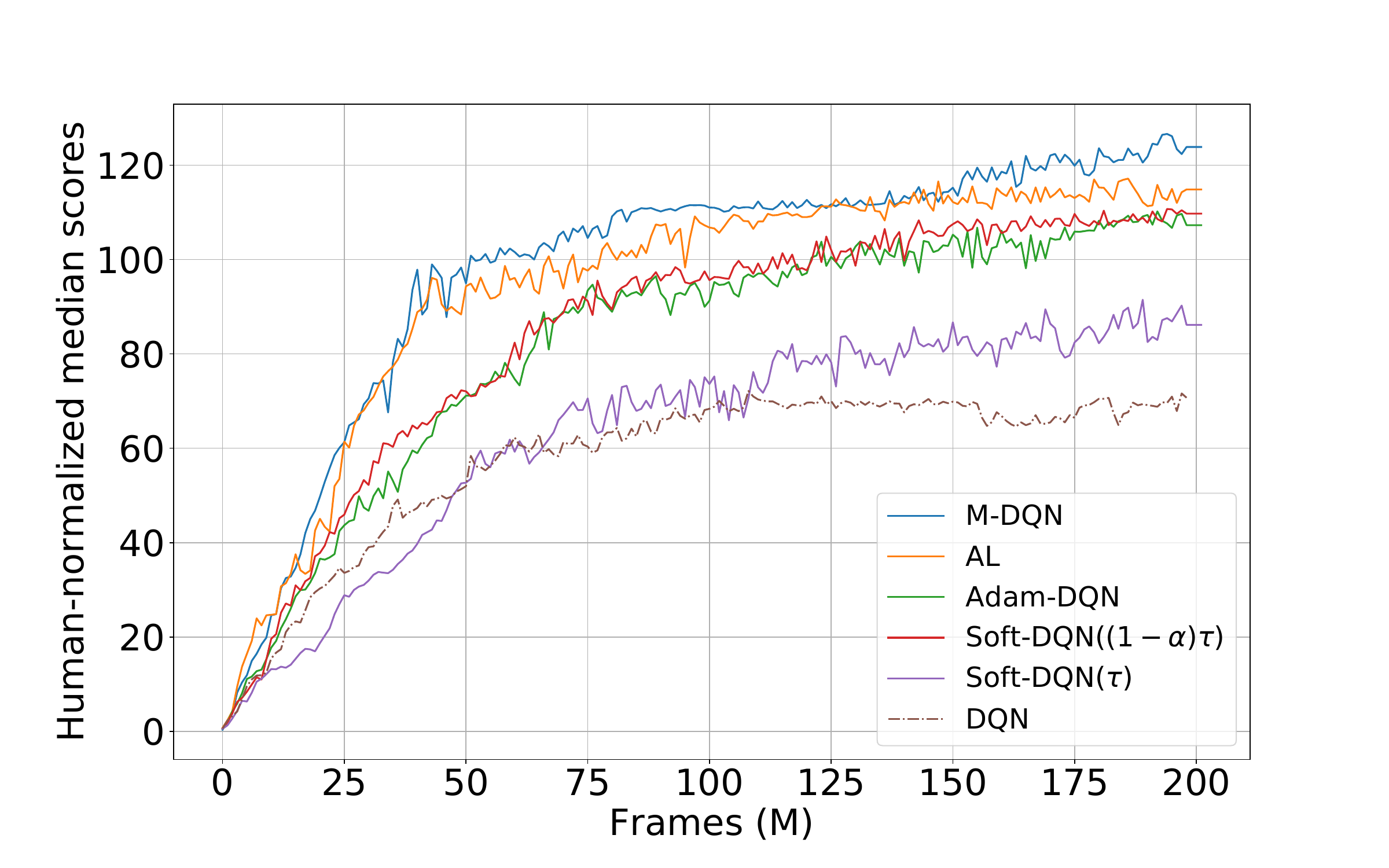}
    \caption{Ablation study of M-DQN: Human-normalized mean (\textbf{left}) and median (\textbf{right}) scores.}
    \label{fig:ablation}
\end{figure}

\begin{figure}
     \centering
     \includegraphics[width=\linewidth]{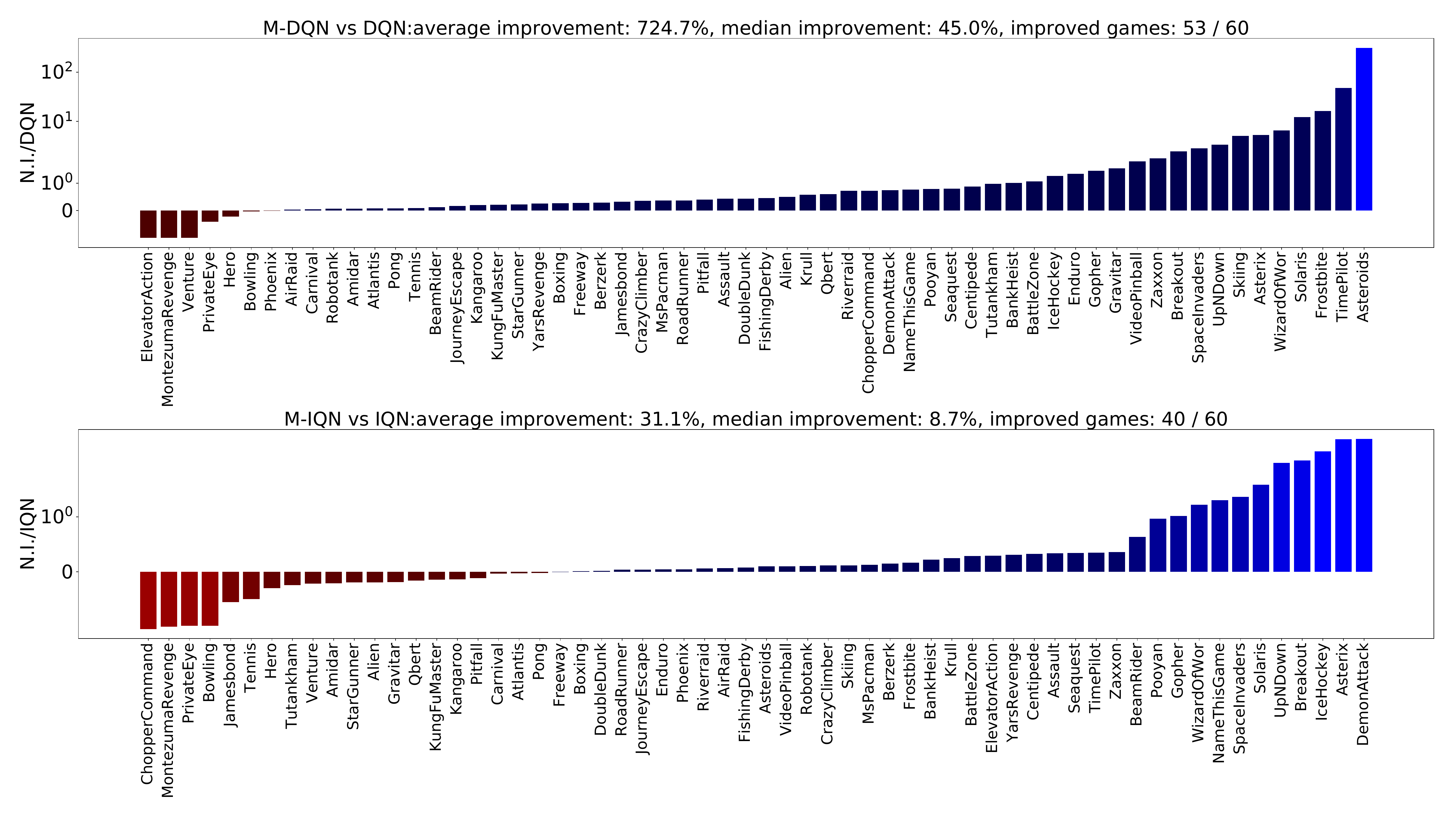}
    \vspace{-4pt}
     \caption{Per-game improvement of M-DQN vs DQN (\textbf{top}) and of M-IQN vs IQN (\textbf{bottom}).}
     \label{fig:aucs}
 \end{figure}

\paragraph{Ablation study.} \label{subsec:ablatioen}
We've build M-DQN from DQN by adding the Adam optimizer (Adam DQN), extending it to maximum entropy RL (Soft-DQN, Eq.~\eqref{eq:soft_dqn_target}), and then adding the Munchausen term (M-DQN, Eq.~\eqref{eq:munchausen_dqn_target}). A natural ablation is to remove the Munchausen term, and use only maximum entropy RL, by considering M-DQN with $\alpha=0$ (instead of $0.9$ for M-DQN), and the same $\tau$ (here, $3e-2$), which would give Soft-DQN($\tau$). However, Thm.~\ref{thm:equivalence} states that M-DQN performs entropy regularization with an implicit coefficient of $(1-\alpha)\tau$, so to compare M-DQN and Soft-DQN fairly, one should evaluate Soft-DQN with such a temperature, that is $3e-3$ in this case. We denote this ablation as Soft-DQN$((1-\alpha)\tau)$.
As sketched in Sec.~\ref{sec:what}, AL can also be seen as a limit case (on an abstract way, as $\tau\rightarrow 0$, see also Appx.~\ref{subappx:m-agent} for details on the algorithm). 
We provide an ablation study of all these variations, all using Adam (except DQN), in Fig.~\ref{fig:ablation}.
All methods perform better than DQN. Adam DQN performs very well and is even competitive with C51. This is an interesting insight, as changing the optimizer compared to the published parameters dramatically improves the performance, and Adam DQN could be considered as a better baseline\footnote{To be on par with the literature, we keep using the published DQN as the baseline for other experiments.}. Surprisingly, if better than DQN, Soft-DQN does not perform better than Adam DQN. This suggests that maximum entropy RL alone might not be sufficient. We kept the temperature $\tau=0.03$, and one could argue that it was not tuned for Soft DQN, but it is on par with the temperature of similar algorithms~\cite{song2018revisiting,vieillard2020leverage}. We observe that AL performs better than Adam DQN. Again, we kept $\alpha=0.9$, but this is consistent with the best performing parameter of~\citet[\textit{e.g.},~Fig.~7]{bellemare2016increasing}. The proposed M-DQN outperforms all other methods, both in mean and median, and especially Soft-DQN by a significant margin (the sole difference being the Munchausen term).

\paragraph{Comparison to the baselines.}

We report aggregated results as Human-normalized mean and median scores on Figure~\ref{fig:human_normalized}, that compares the Munchausen agents to the baselines.
M-DQN is largely over DQN, and outperforms C51 both in mean and median. It is remarkable that M-DQN, justified by theoretically sound RL principles and without using common deep RL tricks like $n$-steps returns, PER or distRL, is competitive with distRL methods. It is even close to IQN (in median), considered as the best distRL-based agent. We observe that M-IQN, that combines IQN with Munchausen principle, is better than all other baselines, by a significant margin in mean.
We also report the final Human-normalized and Rainbow-normalized scores of all the algorithms in Table~\ref{tab:score_summary}.
These results are on par with the Human-normalized scores of Fig.~\ref{fig:human_normalized} (see Appx.~\ref{subappx:metrics} for results over frames). M-DQN is still close to IQN i median, is better than DQN, and C51, while M-IQN is the best agent w.r.t. all metrics.

 \begin{table}[]
     \centering
     \caption{Mean/median Human/Rainbow-normalized scores at $200$M frames, on the 60 games, averaged over $3$ random seeds. In \textbf{bold} are the best of each column, and in \textcolor{blue}{blue} over Rainbow.
     We also provide the number of improved games (compared to Human and Rainbow).
     }%
     \begin{tabular}{l r r r r r r}
\toprule
& \multicolumn{3}{c}{Human-normalized} & \multicolumn{3}{c}{Rainbow-normalized}\\
\cmidrule(r){2-4} \cmidrule(r){5-7}
         & Mean & Median & \#Improved & Mean & Median & \#Improved\\
\midrule
M-DQN & 340\% & 124\% & 37 & 89\% & 92\%& 21 \\
M-IQN & \textbf{\textcolor{blue}{563\%}} & \textbf{\textcolor{blue}{165\%}} & \textbf{43} &\textbf{\textcolor{blue}{130\%}} & \textbf{\textcolor{blue}{109\%}} & \textbf{38}\\ \hdashline
RAINBOW & 414\% & 150\% & \textbf{43} &100\% &100\% & - \\
IQN & \textcolor{blue}{441\%} & 139\% & 41 &\textcolor{blue}{105\%} & 99\% & 27\\
C51 & 339\% & 111\% & 33 & 84\% & 70\% & 11 \\
DQN & 228\% & 71\% & 23 & 51\%  & 51\% & 3\\
\bottomrule
     \end{tabular}
     \label{tab:score_summary}
 \end{table}

\paragraph{Per-game improvements.}
In Figure~\ref{fig:aucs}, we report the improvement for each game of  the Munchausen agents over the algorithms they modify. The ``Munchausened'' versions show significant improvements, on a large majority of Atari games ($53/60$ for M-DQN vs DQN, $40/60$ for M-IQN vs IQN). This result also explains the sometime large difference between the mean and median metrics, as some games benefit from a particularly large improvement.
All learning curves are in Appx~\ref{subappx:metrics}.

\section{Conclusion}
In this work, we presented a simple extension to RL algorithms: Munchausen RL. This method augments the immediate rewards by the scaled logarithm of the policy computed by an RL agent. We applied this method to a simple variation of DQN, Soft-DQN, resulting in the M-DQN algorithm. M-DQN shows large performance improvements: it outperforms DQN on 53 of the 60 Atari games, while simply using a modification of the DQN loss. In addition, it outperforms the seminal distributional RL algorithm C51. We also extended the Munchausen idea to distributional RL, showing that it could be successfully combined with IQN to outperform the Rainbow baseline. Munchausen-DQN relies on theoretical foundations.  To show that, we have studied an abstract Munchausen Value Iteration scheme and shown that it implicitly performs KL regularization. Notably, the strong theoretical results of~\cite{vieillard2020leverage} apply to M-DQN. By rewriting it in an equivalent ADP form, we have related our approach to the literature, notably to CVI, DPP and AL . We have shown that M-VI increases the action-gap, and we have quantified this increase, that can be infinite in the limit. In the end, this  work highlights that a thoughtful revisiting of the core components of reinforcement learning can lead to new and efficient deep RL algorithms.

\clearpage

\bibliographystyle{plainnat}
\bibliography{biblio}
\clearpage

\appendix

\paragraph{Content.} These appendices provide the following additional material:
\begin{itemize}
    \item Appx.~\ref{appx:proofs} details the derivations made in the paper and proves the stated results.
    \item Appx.~\ref{appx:exp} provides additional experimental details, such as a detailed description of the Munchausen agents, and additional results and visualisations.
\end{itemize}

\paragraph{Code.} All the code used for the experiments is available online at \url{https://github.com/google-research/google-research/tree/master/munchausen_rl}.

\section{Detailed derivation and proofs}
\label{appx:proofs}

This appendix provides additional details regarding the derivation sketched in the main paper as well as the proofs of the stated results:
\begin{itemize}
    \item Appx.~\ref{subappx:Soft-DQN} details the derivation of Soft-DQN.
    \item Appx.~\ref{subappx:proof_equivalence} proves the result that relates Munchausen VI to Mirror Descent VI.
    \item Appx.~\ref{subappx:bounds} provides and proves component-wise bounds for Munchausen VI, that also apply to Munchausen-DQN.
    \item Appx.~\ref{subappx:related_works} details the derivation that allows linking the proposed Munchausen approach to the literature.
    \item Appx.~\ref{subappx:proof_action_gap} proves the result quantifying the increase of the action-gap.
\end{itemize}
First, we recall the notations introduced in the main paper as well as some useful facts about (regularized) MDPs.

We write $\Delta_X$ the simplex over the finite set $X$ and $Y^X$ the set of applications from $X$ to the set $Y$. An MDP is a tuple $\{\states, \actions, P, r, \gamma\}$, with $\states$ and $\actions$ the state and action spaces (here assumed finite), $P\in\Delta_{\states}^{\states\times\actions}$ the Markovian transition kernel, $r\in\mathbb{R}^{\states\times\actions}$ the reward function, uniformly bounded by $r_\text{max}$, and $\gamma\in(0,1)$ the discount factor. A policy $\pi\in\Delta_\actions^\states$ associates to each state a distribution over actions (a deterministic policy being a special case), and the quality of a policy is quantified by the expected discounted cumulative return, formalized as the state-action value function, $q_\pi(s,a) = \mathbb{E}_\pi[\sum_{t=0}^\infty \gamma^t r(s_t, a_t) | s_0=s, a_0 = a]$, the expectation being over trajectories induced by the policy $\pi$ and the dynamics. 

For $f,g\in\mathbb{R}^{\states\times\actions}$, we define a component-wise dot product $\langle f, g\rangle = (\sum_{a}f(s,a) g(s,a))_s\in\mathbb{R}^\states$. This will be used with $q$-functions and (log-) policies. For $v\in\mathbb{R}^\states$, we have $P v = (\E_{s'|s,a}[v(s')])_{s,a} \in\mathbb{R}^{\states\times\actions}$. We also defined a policy-induced transition kernel $P_\pi$ as $P_\pi q = P \langle \pi, q\rangle$. With this, the Bellman evaluation operator is $T_\pi q = r + \gamma P_\pi q$ and its unique fixed point is $q_\pi$.

An optimal policy satisfies $\pi_*\in\argmax_{\pi} q_\pi$, component-wise, and the associated (unique) optimal value function $q_* = q_{\pi_*}$ satisfies the Bellman equation $q_*(s,a) = r(s,a) + \gamma E_{s'|s,a}[\max_{a'}q_*(s',a')]$. We write the set of greedy policies as $\gr(q) = \argmax_{\pi\in\Delta_\actions^\states}\langle\pi, q\rangle$. We'll also use softmax policies, $\pi=\softmax(q) \Leftrightarrow \pi(a|s) = \frac{\exp q(s,a)}{\sum_{a'} \exp q(s,a')}$.

We'll also make use of the entropy of a policy, $\h(\pi) = -\langle \pi,\ln \pi\rangle$, and of the KL between two policies, $\kl(\pi_1||\pi_2) = \langle \pi_1, \ln\pi_1 - \ln\pi_2\rangle$. An MDP regularized by a scaled entropy $\tau\h(\pi)$, also known as maximum entropy RL, optimizes for the reward $r - \tau \ln \pi$. It has a unique optimal $q$-function $q_*^\tau$ and a unique optimal policy $\pi_*^\tau$, related by $\pi_*^\tau = \softmax(q_*^\tau)$; it is related to the solution of the unregularized MDP by $\|q_*^\tau - q_*\|_\infty \leq \frac{\tau \ln |\actions|}{1-\gamma}$~\cite{geist2019theory}. We also write $q_\pi^{\tau}$ the value function of the policy $\pi$ in this regularized MDP.

Lastly, by classic properties of the Legendre-Fenchel transform~\cite{boyd2004convex,vieillard2020leverage}, we have  $\forall q\in\mathbb{R}^{\states\times\actions}$:
\begin{equation}
    \max_{\pi\in\Delta_\actions^\states} \langle q,\pi \rangle + \tau \h(\pi) = \tau \ln\langle 1, \exp \frac{q}{\tau}\rangle
    = \langle \pi', q\rangle + \tau \h(\pi') \text{ with } \pi' = \softmax(q).
\end{equation}

\subsection{Derivation of Soft-DQN}
\label{subappx:Soft-DQN}

Soft-DQN can be derived from the maximum entropy RL framework. To do so, it is sufficient to follows the derivation that ~\citet{haarnoja2018soft} made for SAC. In our case, the actions being discrete, no approximation is necessary for computing the policy (there is no actor), which gives Soft-DQN.

Alternatively, and equivalently, one can derive Soft-DQN as an approximate VI scheme for an MDP regularized by a scaled entropy. The regularized VI scheme is~\cite{geist2019theory,vieillard2020leverage}:
\begin{equation}
    \begin{cases}
        \pi_{k+1} = \argmax_\pi \langle \pi, q_k\rangle + \tau \h(\pi)
        \\
        q_{k+1} = r + \gamma P(\langle \pi_{k+1}, q_k \rangle + \tau \h(\pi_{k+1})) + \epsilon_{k+1}
    \end{cases}.
\end{equation}
From Legendre-Fenchel, $\pi_{k+1} = \softmax(q_k)$. Using basic calculus, we have
\begin{equation}
    \langle \pi_{k+1}, q_k \rangle + \tau \h(\pi_{k+1}) = \langle \pi_{k+1}, q_k \rangle - \tau \langle\pi_{k+1}, \ln\pi_{k+1}\rangle = \langle \pi_{k+1}, q_k - \tau \ln \pi_{k+1}\rangle. 
\end{equation}
Thus, we can write equivalently the regularized VI scheme as
\begin{equation}
    q_{k+1} = r + \gamma P \langle \pi_{k+1}, q_k - \tau \ln \pi_{k+1}\rangle + \epsilon_{k+1}, \text{ with } \pi_{k+1} = \softmax(q_k),
\end{equation}
which is basically the Soft-DQN target depicted in Eq.~\eqref{eq:soft_dqn_target}.

\subsection{Proof of Thm.~\ref{thm:equivalence}}
\label{subappx:proof_equivalence}

The proof is similar to the one done in the main paper for the case $\alpha=1$. Recall Eq.~\eqref{eq:munchvi}, that gives an iteration of M-VI($\alpha$,$\tau$):
\begin{equation}
\begin{cases}
    \pi_{k+1} = \argmax_{\pi\in\Delta_\actions^\states}\langle\pi, q_k\rangle + \tau\mathcal{H}(\pi)
    \\
    q_{k+1} = r + \alpha\tau\ln\pi_{k+1} + \gamma P\langle\pi_{k+1}, q_k - \tau \ln \pi_{k+1}\rangle  +  \epsilon_{k+1}.
\end{cases}
\end{equation}
Define for any $k\geq 0$ the term $q'_k$ as
\begin{equation}
    q'_{k} = q_{k} - \alpha \tau \ln \pi_{k}.
\end{equation}
By basic calculus, we can rewrite the evaluation step as follows:
\begin{align}
    q_{k+1} &= r + \alpha\tau\ln\pi_{k+1} + \gamma P\langle\pi_{k+1}, q_k - \tau \ln \pi_{k+1}\rangle  +  \epsilon_{k+1}
    \\
    &=  r + \alpha\tau\ln\pi_{k+1} + \gamma P\langle\pi_{k+1}, q_k - \alpha \tau \ln \pi_k + \alpha \tau \ln \pi_k - \tau \ln \pi_{k+1}\rangle  +  \epsilon_{k+1}
    \\ \Leftrightarrow
    q'_{k+1} &=  r  + \gamma P\langle\pi_{k+1}, q'_k  - \alpha \tau \ln \frac{\pi_{k+1}}{\pi_k} - (1-\alpha) \tau \ln \pi_{k+1} \rangle  +  \epsilon_{k+1}
    \\
    &= r + \gamma P\left(\langle\pi_{k+1}, q'_k\rangle  - \alpha \tau \kl(\pi_{k+1}||\pi_k) + (1-\alpha) \tau \h(\pi_{k + 1}) \right)  +  \epsilon_{k+1}.
\end{align}
For the greedy step, we have:
\begin{align}
    \langle\pi, q_k\rangle + \tau\mathcal{H}(\pi) &= \langle \pi, q_k - \tau \ln\pi\rangle
    \\
    &= \langle\pi, q'_k + \alpha \tau \ln \pi_k - \tau \ln \pi\rangle
    \\
    &= \langle\pi, q'_k - \alpha \tau \ln\frac{\pi}{\pi_k} - (1-\alpha)\tau \ln\pi\rangle
    \\
    &= \langle \pi, q'_k \rangle - \alpha \tau \kl(\pi||\pi_k) + (1-\alpha)\tau \h(\pi).
\end{align}
Therefore, we have shown that
\begin{gather}
    \begin{cases}
        \pi_{k+1} = \argmax_{\pi\in\Delta_\actions^\states}\langle\pi, q_k\rangle + \tau\mathcal{H}(\pi)
        \\
        q_{k+1} = r + \alpha\tau\ln\pi_{k+1} + \gamma P\langle\pi_{k+1}, q_k - \tau \ln \pi_{k+1}\rangle  +  \epsilon_{k+1}
    \end{cases}
    \\ \Updownarrow \\
    \begin{cases}
        \pi_{k+1} = \argmax_{\pi\in\Delta_\actions^\states}\langle \pi, q'_k \rangle - \alpha \tau \kl(\pi||\pi_k) + (1-\alpha)\tau \h(\pi)
        \\
        q'_{k+1} = r  + \gamma P\left(\langle\pi_{k+1}, q'_k\rangle  - \alpha \tau \kl(\pi_{k+1}||\pi_k) - (1-\alpha) \tau \h(\pi_{k+1}) \right)  +  \epsilon_{k+1}
    \end{cases}.
\end{gather}
This is exactly the update rule of  MD-VI($\alpha\tau$, $(1-\alpha)\tau$) by~\citet{vieillard2020leverage}. Initialized with the same policy $\pi_0$ and such that $q'_0 = q_0 - \tau \ln \pi_0$, both algorithms will produce the same sequence of policies (for the same sequence of errors). This is enough for \cite[Thm.~1]{vieillard2020leverage} to apply to M-VI(1,$\tau$), producing the same sequence of policies that MD-VI($\tau$,0), the result bounding component-wise $q_* - q_{\pi_k}$ (it only involves the computed policy). This is also enough for \cite[Thm.~2]{vieillard2020leverage} to apply to M-VI($\alpha$,$\tau$), producing the same sequence of policies that MD-VI($\alpha\tau$, $(1-\alpha)\tau$), the result bounding component-wise $q_*^{(1-\alpha)\tau} - q_{\pi_k}$.

\subsection{Component-wise bounds for Munchausen VI}
\label{subappx:bounds}

We state the component-wise bounds for M-VI, announced in Sec.~\ref{sec:what}. We recall that they apply to M-DQN, as explained in Sec.~\ref{sec:what} (by defining to what corresponds $q_k$ and $\epsilon_k$ for M-DQN). 
First, we provide a bound for the case $\alpha=1$.

\begin{cor}
\label{cor:alpha_equal_one}
    Let $(q_k,\pi_k)_{k\geq 0}$ be the sequence of $q$-functions and policies produced by M-VI(1,$\tau$), with $\pi_0$ the uniform policy and $q_0$ such that $\|q_0 - \tau \ln \pi_0\|_\infty\leq \frac{r_\text{max}}{1-\gamma}$. Define
    \begin{align}
        E_k &= -\sum_{j=1}^k \epsilon_j,
        \\
        \text{and }
        A_k^1 &= (I-\gamma P_{\pi_*})^{-1} - (I-\gamma P_{\pi_{k}})^{-1}.
    \end{align}
    Assume that $\|q_k - \tau\ln\pi_{k}\|_\infty \leq \frac{r_\text{max}}{1-\gamma}$. We have that:
    \begin{equation}
        0 \leq q_* - q_{\pi_{k}} \leq \left|A^1_k \frac{E_k}{k}\right| + 
        \frac{4}{(1-\gamma)^2} \frac{r_\text{max}+\tau\ln|\actions|}{k} \un,
    \end{equation}
    with $\un\in\mathbb{R}^{\states\times\actions}$ the vector whose all components are equal to 1.
\end{cor}
\begin{proof}
    Thanks to Thm.~\ref{thm:equivalence}, M-VI(1,$\tau$) produces the same sequence of policies that MD-VI($\lambda'$,$\tau'$) with $\lambda'=\tau$ and $\tau'=0$, and a sequence of $q$-functions related by $q'_k = q_k - \tau \ln \pi_k$ ($q'_k$ being the $q$-functions computed by MD-VI($\lambda'$,$\tau'$)). Thm.~1 of~\citet{vieillard2020leverage} thus readily applies, the assumption $\|q'_k\|_\infty \leq \frac{r_\text{max}}{1-\gamma}$ translating into $\|q_k - \tau\ln\pi_{k}\|_\infty \leq \frac{r_\text{max}}{1-\gamma}$.
\end{proof}

Notice that that the assumption that $\|q_k - \tau\ln\pi_k\|_\infty \leq \frac{r_\text{max}}{1-\gamma}$ is not strong, it can be ensured by clipping the $q_k$-values (see also~\cite[Rk.~1]{vieillard2020leverage}). Without this, a similar bound would still hold, but with a quadratic dependency of the error term to the horizon, instead of a linear one. Notice that the bound in supremum norm provided in Sec.~\ref{sec:what} is a direct corollary of Cor.~\ref{cor:alpha_equal_one}.

Next, we provide a bound for the case $\alpha<1$.
\begin{cor}
    Let $(q_k,\pi_k)_{k\geq 0}$ be the sequence of $q$-functions and policies produced by M-VI($\alpha$,$\tau$), with $\pi_0$ the uniform policy, and with $0\leq\alpha < 1$.
    For the sequence of policies $\pi_0,\dots,\pi_k$, we define
    \begin{equation}
        P_{k:j} = \begin{cases}
            P_{\pi_k} P_{\pi_{k-1}} \dots P_{\pi_j} \text{ if } j\leq k,
            \\
            I \text{ else},
        \end{cases}
    \end{equation}
    with $I\in\mathbb{R}^{(\states\times\actions)\times(\states\times\actions)}$ the identity matrix. We also define
    \begin{align}
        A^2_{k:j} &= P_{\pi_*^{(1-\alpha)\tau}}^{k-j} + (I-\gamma P_{\pi_{k+1}})^{-1} P_{k:j+1}(I-\gamma P_{\pi_j}),
        \text{and }
        E^\alpha_k &= (1-\alpha)\sum_{j=1}^k \alpha^{k-j} \epsilon_j.
    \end{align}
    With these notations, we have
    \begin{equation}
        0 \leq q_*^{(1-\alpha)\tau} - q_{\pi_{k+1}}^{(1-\alpha)\tau} \leq \sum_{j=1}^k \gamma^{k-j}\left|A^2_{k:j} E^\alpha_j\right| + \gamma^k (1 + \frac{1-\alpha}{1-\gamma}) \sum_{j=0}^k \left(\frac{\alpha}{\gamma}\right)^j \frac{r_\text{max}+(1-\alpha)\tau\ln|\actions|}{1-\gamma}\un.
    \end{equation}
\end{cor}
\begin{proof}
    Thanks to Thm.~\ref{thm:equivalence}, M-VI($\alpha$,$\tau$) produces the same sequence of policies that MD-VI($\lambda'$,$\tau'$) with $\lambda'=\alpha\tau$ and $\tau'=(1-\alpha)\tau$, and a sequence of $q$-functions related by $q'_k = q_k - \alpha\tau \ln \pi_k$ ($q'_k$ being the $q$-functions computed by MD-VI($\lambda'$,$\tau'$)). Thm.~2 of~\citet{vieillard2020leverage} thus readily applies, with
    \begin{equation}
        \beta = \frac{\lambda'}{\lambda'+\tau'} = \frac{\alpha\tau}{\alpha \tau + (1-\alpha)\tau} = \alpha,
    \end{equation}
    which gives the stated result.
\end{proof}
We refer to \cite[Sec.~4.2]{vieillard2020leverage} for an extensive discussion of this bound, but we highlight the fact that it still shows a compensation of errors (through a moving average instead of the average of Cor.~\ref{cor:alpha_equal_one}), something that is desirable.

\subsection{Details on related works}
\label{subappx:related_works}

First, we relate M-VI to CVI. Recall Eq.~\eqref{eq:munchvi}:
\begin{equation}
\begin{cases}
    \pi_{k+1} = \argmax_{\pi\in\Delta_\actions^\states}\langle\pi, q_k\rangle + \tau\mathcal{H}(\pi)
    \\
    q_{k+1} = r + \alpha\tau\ln\pi_{k+1} + \gamma P\langle\pi_{k+1}, q_k - \tau \ln \pi_{k+1}\rangle  +  \epsilon_{k+1}.
\end{cases}
\end{equation}
From the Legendre-Fenchel transform, we have that
\begin{equation}
    \pi_{k+1} = \softmax(\frac{q_k}{\tau}) =  \frac{\exp\frac{q_k}{\tau}}{\langle 1,\exp\frac{q_k}{\tau}\rangle} \Leftrightarrow \tau \ln \pi_{k+1} = q_k - \tau \ln \langle 1, \exp\frac{q_k}{\tau}\rangle.
\end{equation}
Injecting this into the evaluation step, we obtain
\begin{align}
    q_{k+1} &= r + \alpha\tau\ln\pi_{k+1} + \gamma P\langle\pi_{k+1}, q_k - \tau \ln \pi_{k+1}\rangle  +  \epsilon_{k+1}
    \\
    &= r + \alpha(q_k - \tau \ln \langle 1, \exp\frac{q_k}{\tau}\rangle) + \gamma P\langle\pi_{k+1}, q_k - (q_k - \tau \ln \langle 1, \exp\frac{q_k}{\tau}\rangle)\rangle  +  \epsilon_{k+1}
    \\
    &= r + \gamma P(\tau \ln\langle 1, \exp\frac{q_k}{\tau}\rangle) + \alpha(q_k - \tau \ln\langle 1, \exp\frac{q_k}{\tau}\rangle) + \epsilon_{k+1},
\end{align}
which is exactly Eq.~\eqref{eq:cvi}, that is a CVI-like update.

It is a classic result that the sum-log-exp tends towards the hard maximum as the temperature goes to zero (this can be also derived from properties of the Legendre-Fenchel transform):
\begin{equation}
    \lim_{\tau\rightarrow 0} \tau \ln \sum_{a} \exp\frac{q_k(s,a)}{\tau} = \max_{a} q_k(s,a).
\end{equation}
Using this, the limit of the previous CVI-like update is
\begin{equation}
    q_{k+1} = r + \gamma P\langle \pi_{k+1},q_k\rangle + \alpha (q_k - \langle \pi_{k+1}, q_k\rangle + \epsilon_{k+1}) \text{ with } \pi_{k+1} \in\gr(q_k),
\end{equation}
where we have used that $\max_a q_k(\cdot, a) = \langle \pi_{k+1}, q_k\rangle$ with $\pi_{k+1}\in\gr(q_k)$. This is exactly Eq.~\eqref{eq:al}.

\subsection{Proof of Thm.~\ref{thm:action_gap}}
\label{subappx:proof_action_gap}

This is indeed a corollary of Thm.~\ref{thm:equivalence}. First, we handle the case $\alpha<1$. From Thm.~\ref{thm:equivalence}, we know that M-VI($\alpha$,$\tau$) produces the same sequence of policies that MD-VI($\alpha\tau$,$(1-\alpha)\tau$). From \cite[Thm.~2]{vieillard2020leverage}, we now that without error $q'_k = q_k -\alpha\tau \ln \pi_k$ (recall that $q'_k$ is the sequence of $q$-functions computed by MD-VI) converges to $q_*^{(1-\alpha)\tau}$ and that $\pi_k$ converges to $\pi_*^{(1-\alpha)\tau}$ (recall that both algorithms produce the same sequence of policies). From this, we can deduce the limit of $q_k$, the sequence of $q$-function produced by Munchausen VI:
\begin{equation}
    \lim_{k\rightarrow\infty} q_k = q_*^{(1-\alpha)\tau} + \alpha \tau \ln \pi_*^{(1-\alpha)\tau}.
\end{equation}
From basic properties of regularized MDPs~\cite{geist2019theory}, we know that
\begin{equation}
    \pi_*^{(1-\alpha)\tau} = \softmax(\frac{q_*^{(1-\alpha)\tau}}{(1-\alpha)\tau})
    \Leftrightarrow
    \ln \pi_*^{(1-\alpha)\tau} = \frac{q_*^{(1-\alpha)\tau}}{(1-\alpha)\tau} - \ln\langle 1,\exp\frac{q_*^{(1-\alpha)\tau}}{(1-\alpha)\tau}\rangle.
\end{equation}
Therefore, we have that
\begin{align}
    \lim_{k\rightarrow\infty} q_k &= q_*^{(1-\alpha)\tau} + \alpha \tau \ln \pi_*^{(1-\alpha)\tau}
    \\
    &= q_*^{(1-\alpha)\tau} + \alpha\tau \left(\frac{q_*^{(1-\alpha)\tau}}{(1-\alpha)\tau} - \ln\langle 1,\exp\frac{q_*^{(1-\alpha)\tau}}{(1-\alpha)\tau}\rangle\right)
    \\
    &= \frac{1+\alpha}{1-\alpha} q_*^{(1-\alpha)\tau} - \frac{\alpha\tau}{1-\alpha} \ln\langle 1,\exp\frac{q_*^{(1-\alpha)\tau}}{(1-\alpha)\tau}\rangle.
\end{align}
Noticing that the log-sum-exp does not depend on the actions, we obtain the stated result.

Next, we handle the case $\alpha=1$. From Thm.~\ref{thm:equivalence}, we know that M-VI(1,$\tau$) produces the same sequence of policies that MD-VI($\tau$,$0$). From \cite[Thm.~1]{vieillard2020leverage}, we now that without error $q'_k = q_k -\alpha\tau \ln \pi_k$ converges to $q_*$ and that $\pi_k$ converges to $\pi_*$, the solutions of the unregularized MDP. To simplify and without much loss of generality, assume that this MDP admits a unique optimal policy. As $q_k = q'_k + \alpha \ln \pi_k$, taking the limit we get for any $s\in\states$
\begin{equation}
    \lim_{k\rightarrow\infty} q_k(s,a) = \begin{cases}
        q_*(s,a) \text{ if } \pi_*(a|s) = 1
        \\
        -\infty \text{ else}
    \end{cases}.
\end{equation}
With the adopted convention, this proves the result for the case $\alpha=1$.

\section{Additional experimental details and results}
\label{appx:exp}

This appendix provides a complete description of the Munchausen agents, it gives additional experimental details, and it proposes additional results and visualisations:
\begin{itemize}
    \item Appx.~\ref{subappx:m-agent} provides a complete description of the Munchausen agents, as well as some additional details for the considered metrics (such as human scores for games not reported in the literature) and for the learning setting.
    \item Appx.~\ref{subappx:comparison} discusses the difference between playing $\varepsilon$-greedy and stochastic policies for Munchausen DQN.
    \item Appx.~\ref{subappx:n_steps} discusses the diffrence between using $1$-step or $3$-steps returns in M-IQN.
    \item Appx.~\ref{subappx:settings} provides elements of comparison with the original ALE setting.
    \item Appx.~\ref{subappx:ablation} provides complementary results for the ablation study.
    \item Appx.~\ref{subappx:metrics} provides complementary comparison results.
\end{itemize}

\subsection{Detailed description of the Munchausen agents}
\label{subappx:m-agent}
All the agents follow a similar learning procedure, described as a pseudo-code in Alg.~\ref{algo:m-dqn} for M-DQN. What changes is the loss that is optimized.

\paragraph{M-DQN.} Here, we  recall the basic workings of M-DQN. It estimates a $q$-value through an online $q$-network $q_\theta$ of weights $\theta$. Every $C$ steps, the weights are copied to a \emph{target} network $q_{\bar\theta}$ of weights $\bar\theta$. Transitions $(s_t, a_t, r_t, s_{t+1})$ are stored in fixed-sized FIFO replay buffer. To collect them, M-DQN interacts with the environment using the policy $\gr_{\varepsilon}(\theta)$, the policy that is $\varepsilon$-greedy with respect to $q_\theta$. M-DQN uses (as DQN) a decay on $\varepsilon$ to favour exploration in the beginning of the learning. Each $F$ steps, M-DQN samples a random batch  $B$ of transitions from $\mathcal{B}$ and minimizes the following loss, based on the regression target of Eq.~\eqref{eq:munchausen_dqn_target}:
\begin{align}
\label{eq:mdqn_loss}
    &\mathcal{L}_{\text{m-dqn}}(\theta) =
    \\
    &\hat{\mathbb{E}}_B \Bigg[h\Big(r_t + \alpha\left[\tau\ln\pi_{\bar\theta}(a_t|s_t)\right]_{l_0}^0 + \gamma\sum_{a\in\actions}\pi_{\bar\theta}(a|s_{t+1})\left(q_{\bar\theta}(s_{t+1},a) - \tau\ln\pi_{\bar\theta}(a|s_{t+1})\right) 
    -  q_\theta(s_t,a_t)\Big)\Bigg],
\end{align}
with $\pi_{\bar\theta} = \softmax(\frac{q_{\bar\theta}}{\tau})$ and $h$ the Huber loss function, with a paremeter $x_h$, $h(x) = x^2$ if   $x < x_h$ else $\abs{x}$. A pseudo-code detailing the learning procedure is given in Alg.~\ref{algo:m-dqn}. 

\begin{algorithm}[tbh]
\caption{Munchausen DQN }
\begin{algorithmic}%
\label{algo:m-dqn}
\REQUIRE $T\in \mathbb{N^*}$ the number of environment steps, $C\in \mathbb{N^*}$ the update period, $F \in \mathbb{N^*}$ the interaction period.
\STATE Initialize $\theta$ at random
\STATE $\mathcal{B} = \{\}$
\STATE $\bar\theta = \theta$
\FOR{$t = 1$ \TO $T$}
    \STATE Collect a transition $b = (s_t, a_t, r_t, s_{t+1})$ from $\gr_e(\theta)$
    \STATE $\mathcal{B} \leftarrow \mathcal{B} \cup \{b\}$
    \IF{$t \mod F == 0$}
        \STATE On a random batch of transitions $B_{t} \subset \mathcal{B}$, update $\theta$ with one step of SGD on $\mathcal{L}_{\text{m-dqn}}$, see~\eqref{eq:mdqn_loss} 
    \ENDIF
    \IF{$k \mod C == 0$}
        \STATE $\bar\theta \leftarrow \theta$
    \ENDIF
\ENDFOR
\RETURN $\gr_0(\theta)$
\end{algorithmic}
\end{algorithm}

\paragraph{AL.} We have shown in Sec.~\ref{sec:what} that AL can be seen as a limiting case of M-DQN, in the limit $\tau\rightarrow 0$. Yet, it cannot be obtained simply by setting $\tau=0$ in Alg.~\ref{algo:m-dqn}. Instead, we rewrite the minimized loss, according to Sec.~\ref{sec:what}. Each $F$ steps, AL samples a random batch  $B$ of transitions from $\mathcal{B}$ and minimizes the loss
\begin{equation}
    \mathcal{L}_{\text{al}}(\theta) = \hat{\mathbb{E}}_B \left[h \left(r_t + \alpha\left(q_{\bar\theta}(s_t,a_t) - \max_{a\in\actions} q_{\bar\theta}(s_t,a) \right)+ \max_{a_\in\actions}q_{\bar\theta}(s_{t+1},a)  -  q_\theta(s_t,a_t)\right)\right].
\end{equation}

\paragraph{M-IQN.} IQN is a distributional method. It does not estimate directly a $q$-function, but the distribution of the discounted cumulative rewards, a so-called $z$-function. Precisely, the $z$-function $z_\pi \in \mathbb{R}^{\states\times\actions}$ of a policy $\pi$  is a random quantity defined, for each $s,a \in \states\times\actions$ as:
\begin{equation}
  z_\pi(s,a) = \sum_{t=0}^\infty\gamma^tr(s_t,a_t), \text{ with }  a_t \sim \pi(\cdot|s_t) \text{ and } s_{t+1} \sim P(\cdot|s_t, a_t) \text{ for } s_0=s \text{ and } a_0=a.
\end{equation}
The $q$-function can be directly related to it with
\begin{equation}
  q_\pi(s,a) = \mathbb{E}\left[z_\pi(s,a)\right].
\end{equation}

A remarkable result is that $z_\pi$ satisfies a Bellman equation, similarly to $q_\pi$, and thus can be estimated with TD. Here, we give a quick  overview of IQN, and explain how we modified it. We refer to~\citet{dabney2018implicit} for an exact derivation and more details of the original algorithm. IQN estimates the quantile function of $z$ at $\sigma\in [0,1]$, denoted $z_\sigma$. The estimated $q$-value is then $\tilde{q}(s,a) = \mathbb{E}_{\sigma\sim U_{[0,1]}}[z_\sigma(s,a)]$, this expectation being practically approximated by Monte Carlo. The TD error of IQN at step $t$, defined with $\sigma, \sigma' ~\sim U_{[0,1]}$, is:
\begin{equation}
    \text{TD}_{\text{IQN}} = r_t + \gamma z_{\sigma'}(s_{t+1}, \pi(s_{t+1})) - z_{\sigma}(s_t,a_t), \text{ with } \pi(s) = \argmax_{a\in\actions}\tilde{q}(s,a).
\end{equation}
In practice, $z_{\sigma'}$ is given by a target network, and $z_\sigma$ by an online network, to be optimized.
The loss is then estimated as the empirical mean of the TD errors, by sampling $\sigma$ and $\sigma'$ uniformly in $[0,1]$. In M-IQN, we use an additional Munchausen term in TD error,
\begin{equation}
    \text{TD}_{\text{M-IQN}} = r_t + \alpha\left[\tau\ln\pi(a_t | s_t)\right]_{l_0}^0 + \gamma\sum_{a\in\actions} \pi(a|s_{t+1})(z_{\sigma'}(s_{t+1}, a) - \tau\ln\pi(a|s_{t+1})) - z_{\sigma}(s_t,a_t)
\end{equation}
with $\pi(\cdot|s) = \softmax(\frac{\tilde{q}(s,\cdot)}{\tau})$ (that is, the policy is softmax with $\tilde{q}$, the quantity with respect to which the original policy of IQN is greedy). We use the same parametrization for $z$ as~\citet{dabney2018implicit}, and all their provided hyperparameters, as implemented in Dopamine. We used the ``Munchausen-RL parameters'' from Table~\ref{tab:hypers}.

\paragraph{Custom log-sum-exp trick.}
Eq.~\ref{eq:mdqn_loss} relies on computing a log-policy, so in our case the log-softmax of a $q$-values. Such computations are usually done using the ``log-sum-exp trick'', that allows for numerically stable operations by factorizing a maximum. This trick is widely used in software libraries, for example in TensorFlow~\cite{tensorflow2015-whitepaper}, used to implement the experiments of this work. With this approach, we use the fact that
\begin{equation}
 \tau \ln\pi_{k+1} = q_k  - \tau \ln \langle 1, \exp \frac{q_k}{\tau} \rangle,
\end{equation}
that can be unstable if $\tau$ is small. Thus, we compute the log-policy terms using a log-sum-exp-trick as
\begin{equation}
    \tau \ln\pi_{k+1} = q_k - v_k  - \tau \ln \langle 1, \exp \frac{q_k - v_k}{\tau} \rangle,
\end{equation}
where we defined $v_k \in \mathbb{R}^\states$ as $v_k(s) = \max_a q_k(s,a)$. This is more stable than the one implemented by default, because it takes into account the temperature coefficient.

\paragraph{Parameters.}
We provide the hyperparameters used in our algorithms in Table~\ref{tab:hypers}. We denote neural networks structures as follow: $\Conv_{a,b}^d c$ is a 2D convolutional layer with $c$ filters of size $a\times b$ and of stride $d$, and $\FC n$ is a fully convolutional layer with $n$ neurons. The parameters of the baseline agents are those reported in Dopamine (with the slight modification of considering $1$-step returns instead of $n$-step returns for IQN, to match the original paper and the algorithm we modify).

\begin{table}[tbh]
    \centering
    \caption{Parameters used for Munchausen RL agents.}
    \begin{tabular}{l r}
    \toprule
    Parameter     & Value \\
    \midrule
    \multicolumn{2}{c}{Base (Adam) DQN parameters} \\
    \midrule
    $C$ (update period)    & 8000\\
    $F$ (interaction  period)    & 4\\
    $\gamma$ (discount) & 0.99\\
    $|\mathcal{B}|$ (replay buffer size) & $10^6$\\
    $|B_{t}|$ (batch size) & 32 \\
    $e_t$ (random actions rate) & 0.01 (with a linear decay of period $2.5\cdot10^5$ steps)\\
    $Q$-network structure & $\Conv_{8,8}^{4}32-\Conv_{4,4}^{2}64-\Conv_{3,3}^{1}64-\FC512-\FC n_A$\\
    activations & Relu\\
    optimizer & Adam ($lr=5e-5$) \\
    \midrule
    \multicolumn{2}{c}{Munchausen-RL specific parameters} \\
    \midrule
    $\tau$ (entropy temperature) & 0.03\\
    $\alpha$ (Munchausen scaling term) & 0.9 \\
    $l_0$ (clipping value) & -1\\
    \midrule
    \multicolumn{2}{c}{AL specific parameters} \\
    \midrule
    $\alpha$ (advantage scaling term) & 0.9 \\
    \bottomrule
    \end{tabular}
    \label{tab:hypers}
\end{table}

\paragraph{Environment details.} We follow the procedures of ~\citet{machado2018revisiting} to train on the ALE. Notably, we perform one training step (a gradient descent step) every $4$ frames encountered in the environment. The state of an agent is the concatenation of the last $4$ frames, sub-sampled to a shape of ($84$, $84$), in gray levels. We refer to~\citet{machado2018revisiting} for details on the preprocessing.

\paragraph{Metrics.} Here, we recall the definitions of the metrics used to compare algorithms. As an aggregating metric, we use the baseline-normalized score. Every $1M$ frames, we compute the undiscounted return averaged over the last $100$ episodes $a_k$, then we normalized it by a random score $r$ and a baseline score $b$ (score after training for 200M steps). The normalized score is then $\frac{a_k - r}{|b - r|}$.  We also use human-normalized scores, when we replace the baseline score by the score of a human. We used human scores reported by~\cite{mnih2015human}. For AirRaid, Carnival, ElevatorAction, JourneyEscape, and Pooyan, not considered in~\citet{mnih2015human}, we averaged scores from game-play posted online by players. 
For a game-per-game metric, we compute the normalized improvement according to a basline. The ``final score'' of an agent is defined as the score averaged over the last $5$M frames. The normalized improvement of a final score $a$ w.r.t. the final score of a baseline $b$ is $\frac{a - b}{|b - r|}$.
The maximum scores reported in Table~\ref{tab:score_all} are the maximum scores over training, averaged over $100$ episodes, averaged over $3$ random seeds, obtained during training.

\subsection{Comparison of greedy and stochastic policies}
\label{subappx:comparison}

\begin{figure}
    \centering
    \includegraphics[width=.49\linewidth]{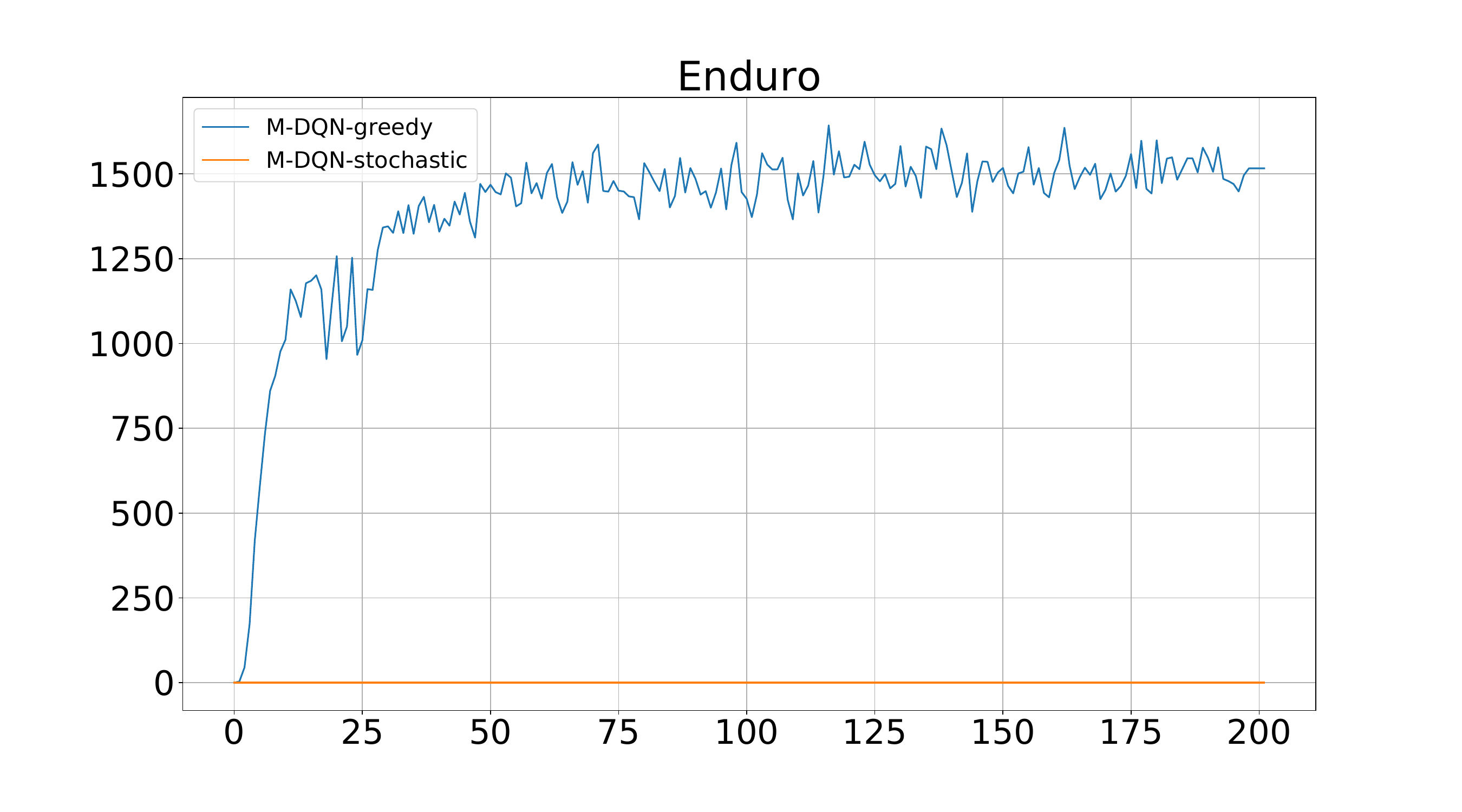}
\includegraphics[width=.49\linewidth]{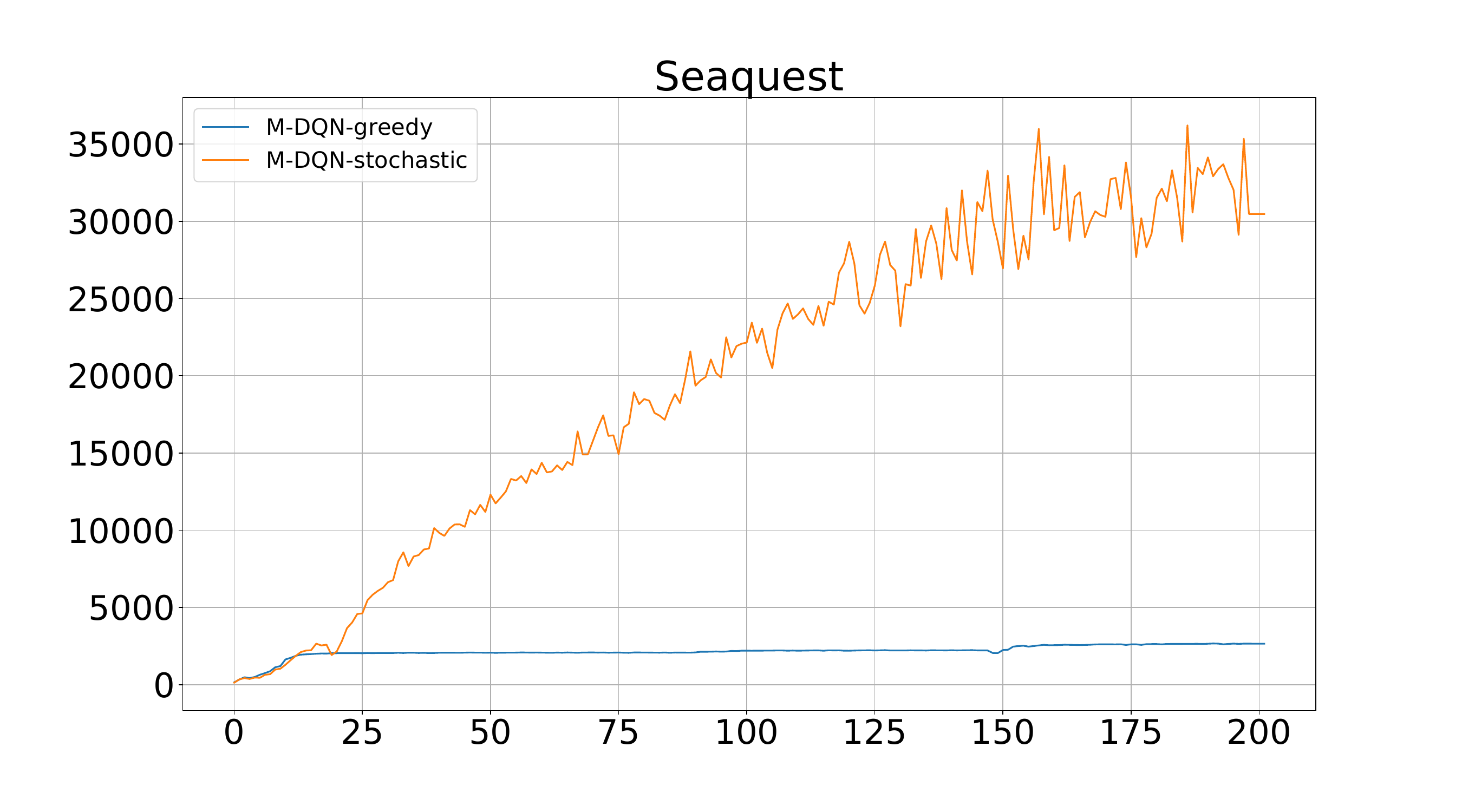}
    \caption{Comparsion of M-DQN with a greedy (blue) or stochastic (orange) interaction policy. \textbf{Left}: Enduro. \textbf{Right:} Seaquest. On Enduro, the stochastic policy is not able to see any reward signal in the beginning, and learns nothing. On Seaquest, we see that it improves over the greedy policy.}
    \label{fig:g_sto_games}
\end{figure}

Although M-DQN naturally produces stochastic policies, we used the $\varepsilon$-greedy one (with respect to $q_\theta)$, as explained in Sec.~\ref{sec:experiments}. This is motivated by the behaviour of some games. In some games, a random policy fails to gather rewards (as for example Venture or Enduro). The $Q$-network is initialized with small $Q$-values, close to zero. Even with the small temperature $\tau=0$ we consider, the resulting softmax policy is very close to uniform, and the M-DQN fails to collect rewards, and thus receives no signal to learn. On the converse, an $\varepsilon$-greedy exploration will have a more (randomly) structured exploration, as the scale of $Q$-values does not matter in this case. It then succeed to gather rewards, and to learn something. This is exemplified in Fig.~\ref{fig:g_sto_games}, left, for the game Enduro.

On the converse, if the agent manage to get rewards, the M-DQN agent with a stochastic policy will perform more exploration, and a directed one, as it will chose more often actions with high $Q$-values, thanks to the softmax policy. Consequently, thanks to this less random exploration, it could perform better. We hypothesize that it is what happens for the game Seaquest, shown in Fig.~\ref{fig:g_sto_games}, right.

\begin{figure}
    \centering
     \includegraphics[width=0.49\linewidth]{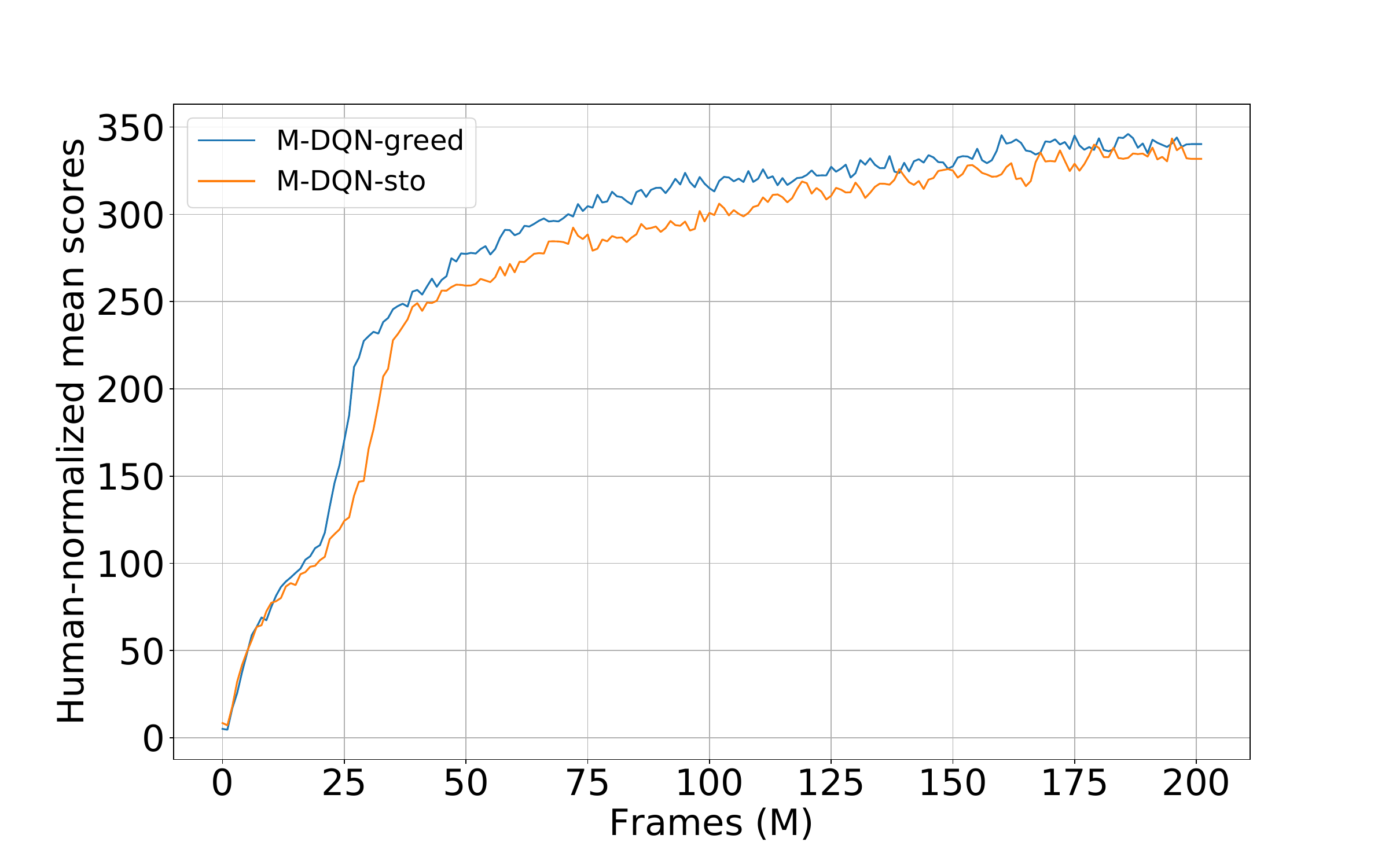}
    \includegraphics[width=0.49\linewidth]{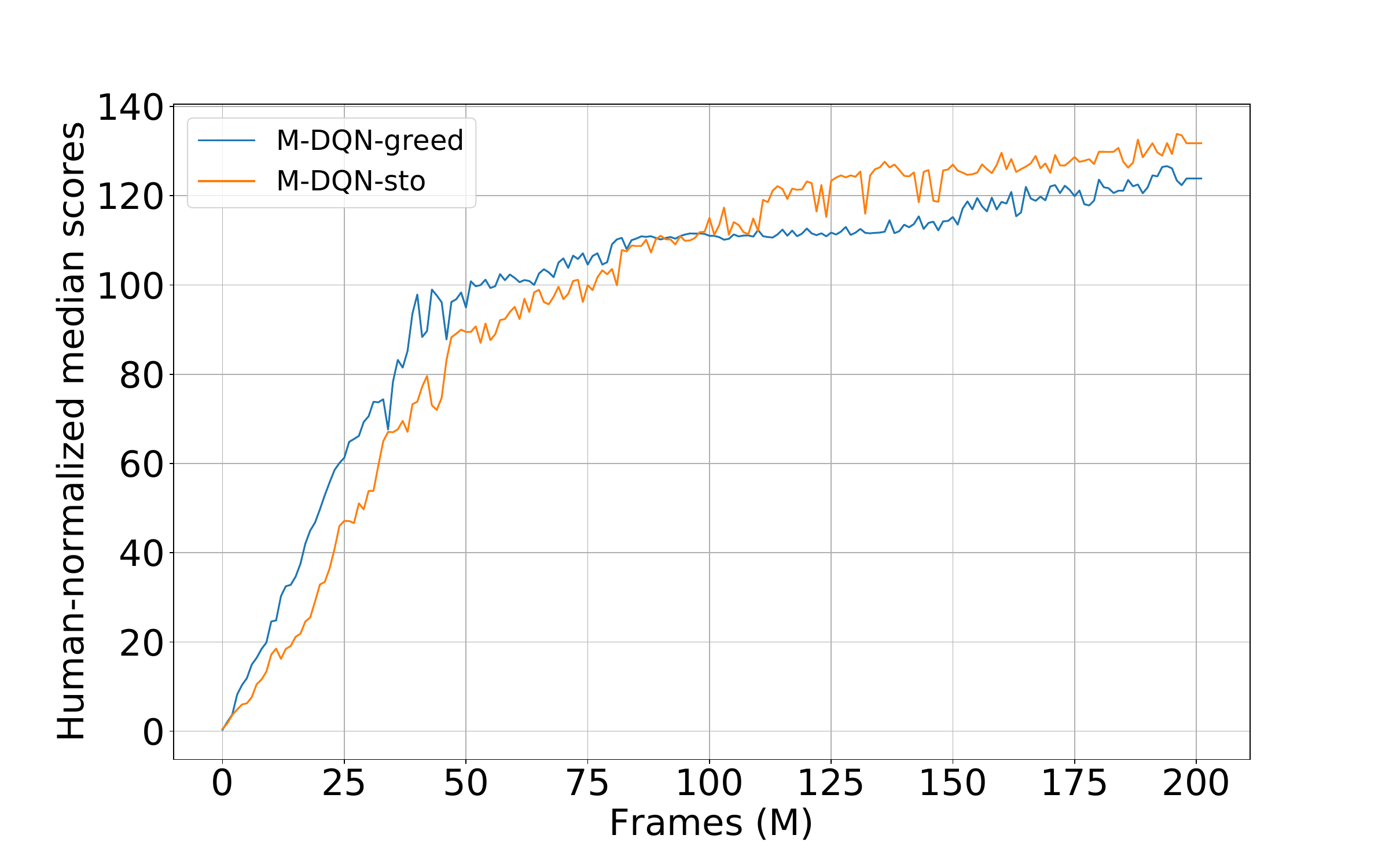}
    \caption{Human-normalized scores of M-DQN greedy and stochastic, mean (\textbf{left}) and median \textbf{right}).}
    \label{fig:gr_sto_human}
\end{figure}

In Fig.~\ref{fig:gr_sto_human}, we provide the Human-normalized scores of both options, playing with an $\varepsilon$-greedy policy or with the more natural stochastic one. We observe that the stochastic policy is slightly better in median. Yet, it improves less games too, and we kept the $\varepsilon$-greedy policy for the core results. Improving the stochastic policy, maybe with an adaptive temperature or an adaptive $\alpha$ parameter, is an interesting future direction of research.

\subsection{Comparison of $1$-step and $3$-steps learning in M-IQN}
\label{subappx:n_steps}
\begin{figure}
    \centering
    \includegraphics[width=.49\linewidth]{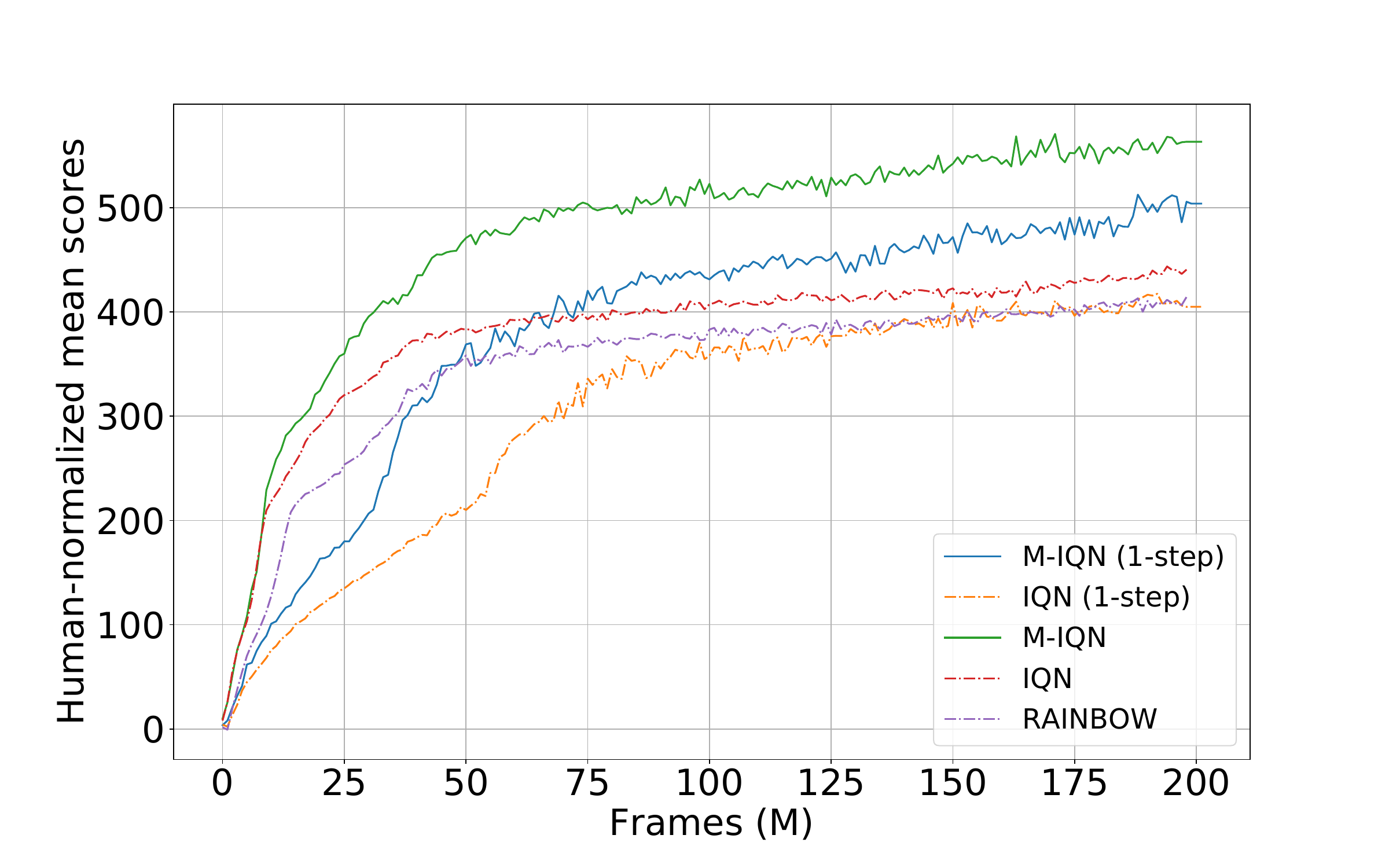}
    \includegraphics[width=.49\linewidth]{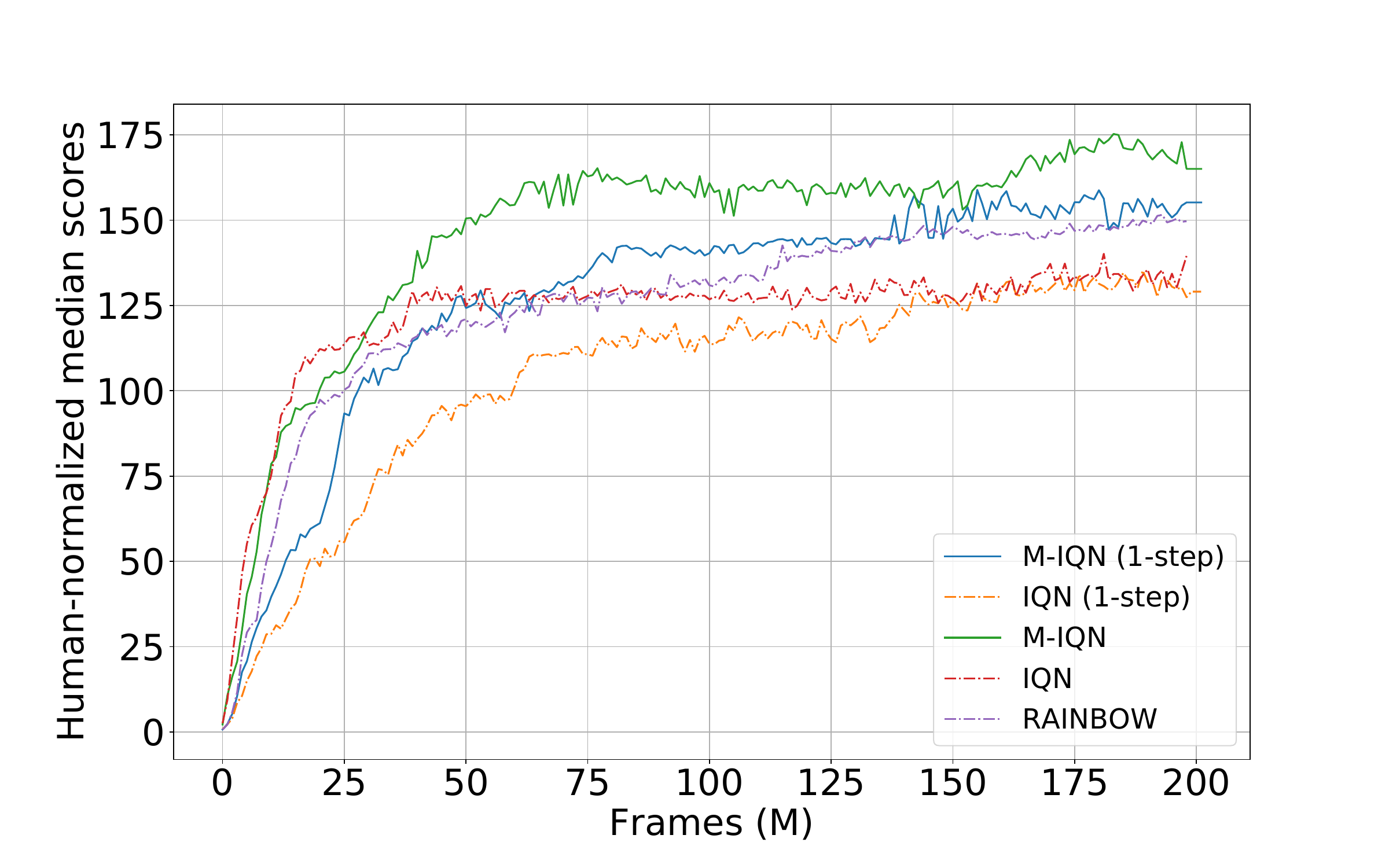}
    \caption{Human normalized scores of M-IQN, IQN, and Rainbow with different $n$-steps returns, mean (\textbf{left}) and median \textbf{right}). M-IQN, IQN, and Rainbow use $3$-steps, while the other two use $1$-step.}
    \label{fig:n_steps}
\end{figure}

The results in the papers are computed with a version of M-IQN that uses $3$-steps learning, and compared to version of IQN that also uses $3$-steps learning (as it is by default in the Dopamine library). For completeness, we evaluate M-IQN with $1$-steps returns, and compre it to IQN with $1$-step returns. The human-normalized scores for these algorithms are reported in Fig.\ref{fig:n_steps}. Theses results show that (1) $n$-step learning and M-RL combine efficiently, as M-IQN $3$-steps clearly outperforms M-IQN $1$-step and (2) that M-IQN alone (with only $1$-step returns) yields already high performances, and it particular outperforms -- although by a tight margin -- the Rainbow baseline, that uses $3$-steps returns.

\subsection{Element of comparison with the original ALE setting}
\label{subappx:settings}

We explained in Sec.~\ref{sec:experiments} the difference between the ALE setting we consider, more modern and more difficult, compared to the ALE setting often considered, for example for the seminal DQN~\cite{mnih2015human} or for Rainbow~\cite{hessel2018rainbow}. The Rainbow baseline we consider~\cite{castro2018dopamine} is also not exactly the published one: even if the most important features are included, as deemed by~\citet{hessel2018rainbow}, it does not include all features (such as double $Q$-learning or dueling architecture).

\begin{figure}
    \centering
    \includegraphics[width=.49\linewidth]{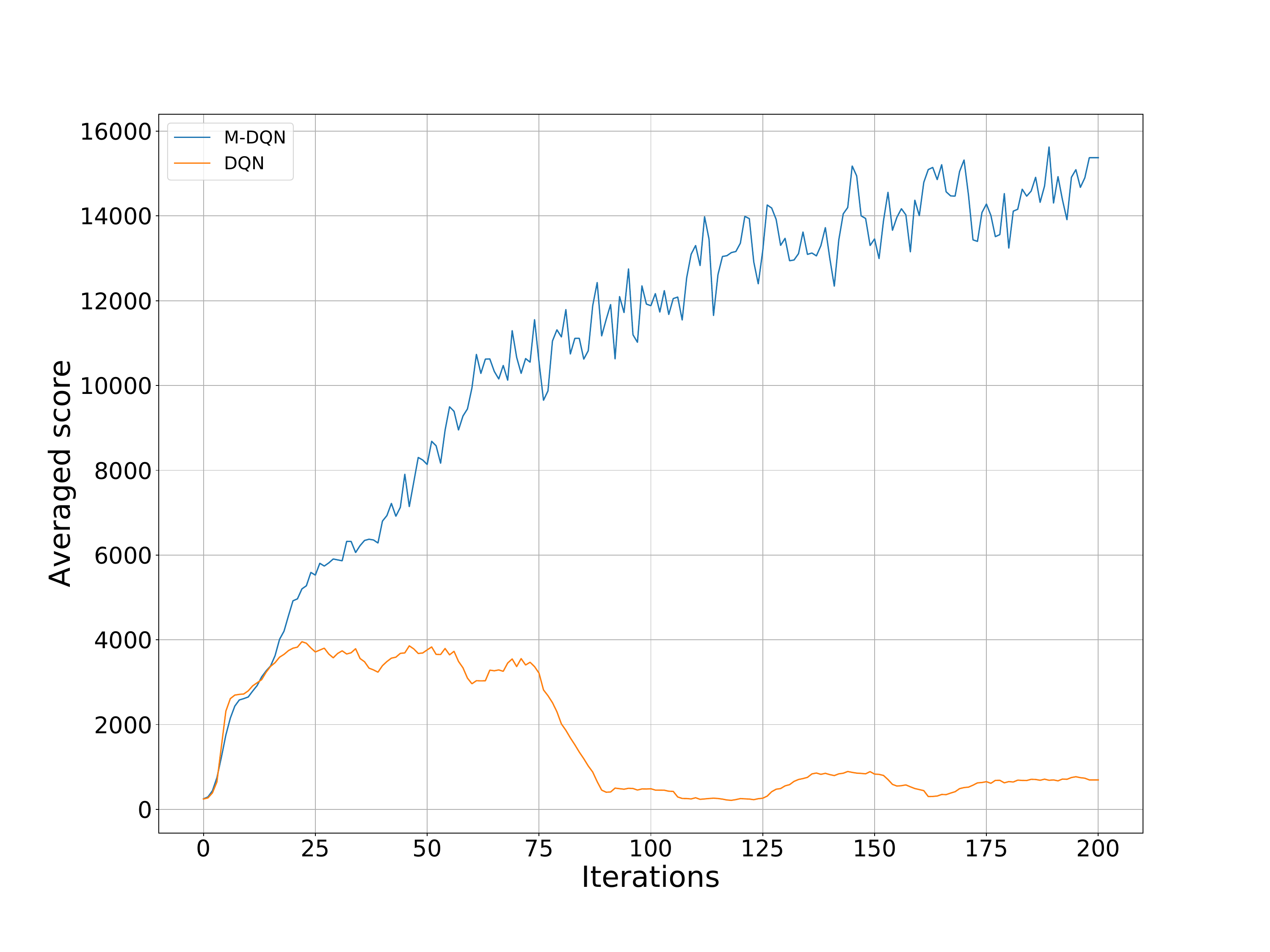}
    \includegraphics[width=.49\linewidth]{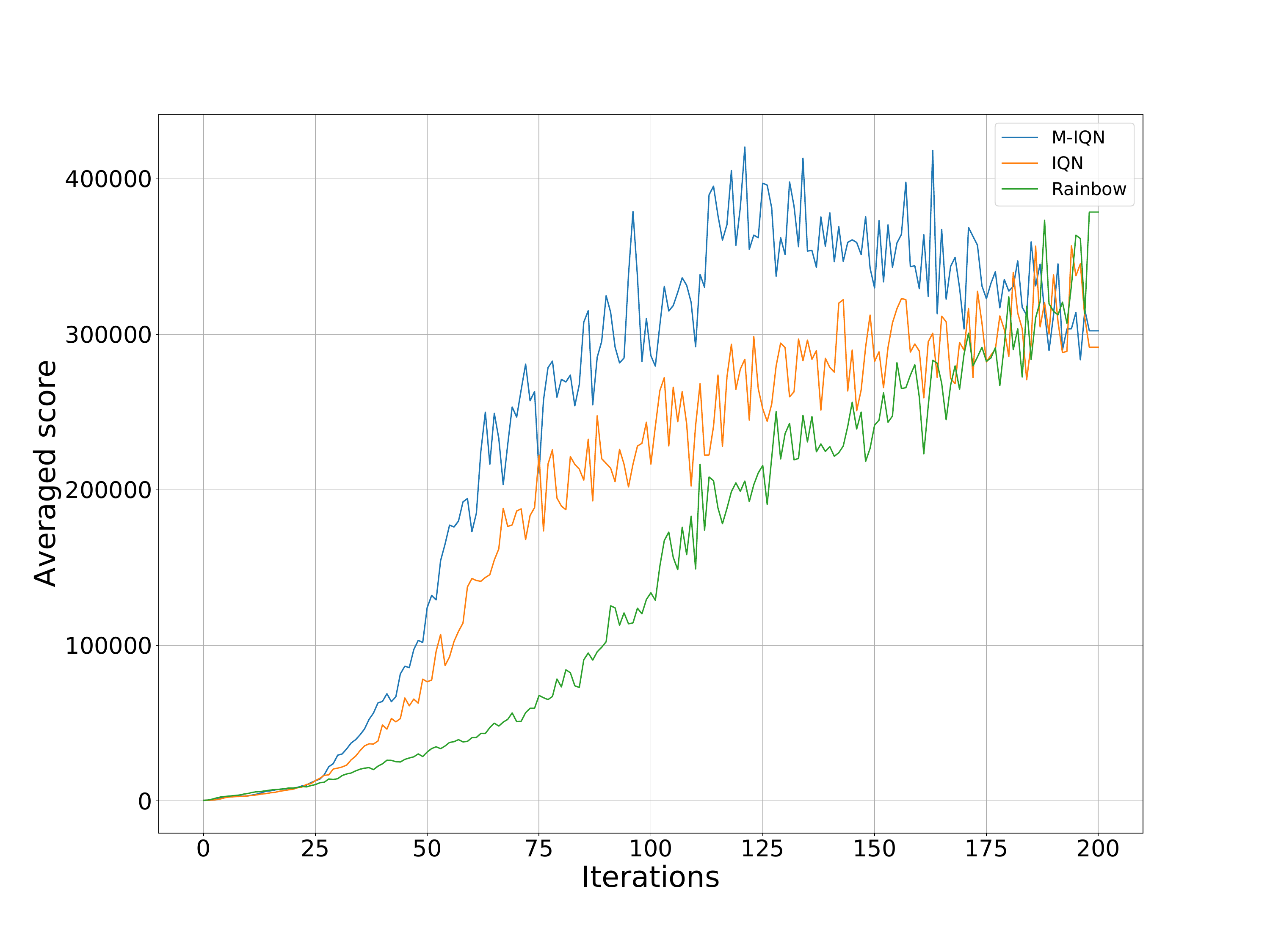}
    \caption{Scores of different agent on the game Asetrix, using the original ALE. \textbf{left:} M-DQN and DQN. \textbf{right:} Rainbow, IQN and M-IQN.}
    \label{fig:asterix_original}
\end{figure}

As a (partial) check, we also evaluated our Munchausen agents, M-DQN and M-IQN, as well as the baselines DQN, IQN and Rainbow, in a setting as close as possible to the one used for the baselines' publications. Notably, here we did not used sticky actions, making the environment deterministic, and we end an episode whenever the agent loses a life, instead of when it encounters a game-over. We also use hyperparameters provided in the original publications, the only difference being that we used a target update period of $10000$ steps instead of $8000$. We did so on the Asterix game, the results being depicted in Fig.~\ref{fig:asterix_original}.

On Fig.~\ref{fig:asterix_original}, left, we can observe DQN and M-DQN. The result for DQN is normal, despite the apparent ``crash'', see for example the training curves in~\cite{hessel2018rainbow} (notice also that it is often the best scores over training which is reported, instead of the final one, as in our Tab.~\ref{tab:score_all} or in the seminal DQN publication~\cite{mnih2015human}). We can observe that M-DQN performs much better than DQN, without falling, and that the score is close to the one of M-DQN in the more difficult setting (15k vs 19k in the more difficult setting).

On Fig.~\ref{fig:asterix_original}, right, we can observe Rainbow, IQN and M-IQN. All algorithms perform pretty well. For example, Rainbows reaches roughly 350k, comparable to the original publication\footnote{The setting is still not exactly the same, due to less enhancements in the Dopamine's Rainbow, a different codebase, but also a difference in the start (human start vs no-op for Rainbow, straight start for us), and possibly a different ROM, which cannot be checked.}. This is much more than in  our setting, where Rainbow reaches only 18k, suggesting that the original setting is easier. We can also see that IQN works well (and somehow surprisingly better than in the original publication, compared to Rainbow), and that M-IQN works better than both IQN and Rainbow.

An interesting thing is to see how the methods degrades (roughly) when going from the agent is trained in the considered setting, compared to the original one. Rainbow goes from 350k to 18k (5\% of the original scores), IQN goes from 350k to 33k (10\%), while M-DQN goes from 15k to 17k (113\%) and M-IQN goes from 350k to 50k (17\%). This suggests that M-RL might be more stable over environments.

For sure, this discussion only holds for one game, and no general conclusion can be drawn. Yet, it suggests a few things, the ALE setting we consider is more difficult, among other advantages~\cite{machado2018revisiting}, the Rainbow baseline we consider is correct, and M-RL seems to be more stable.

\subsection{Additional results on the ablation study}
\label{subappx:ablation}

\begin{figure}
    \centering
    \includegraphics[width=.49\linewidth]{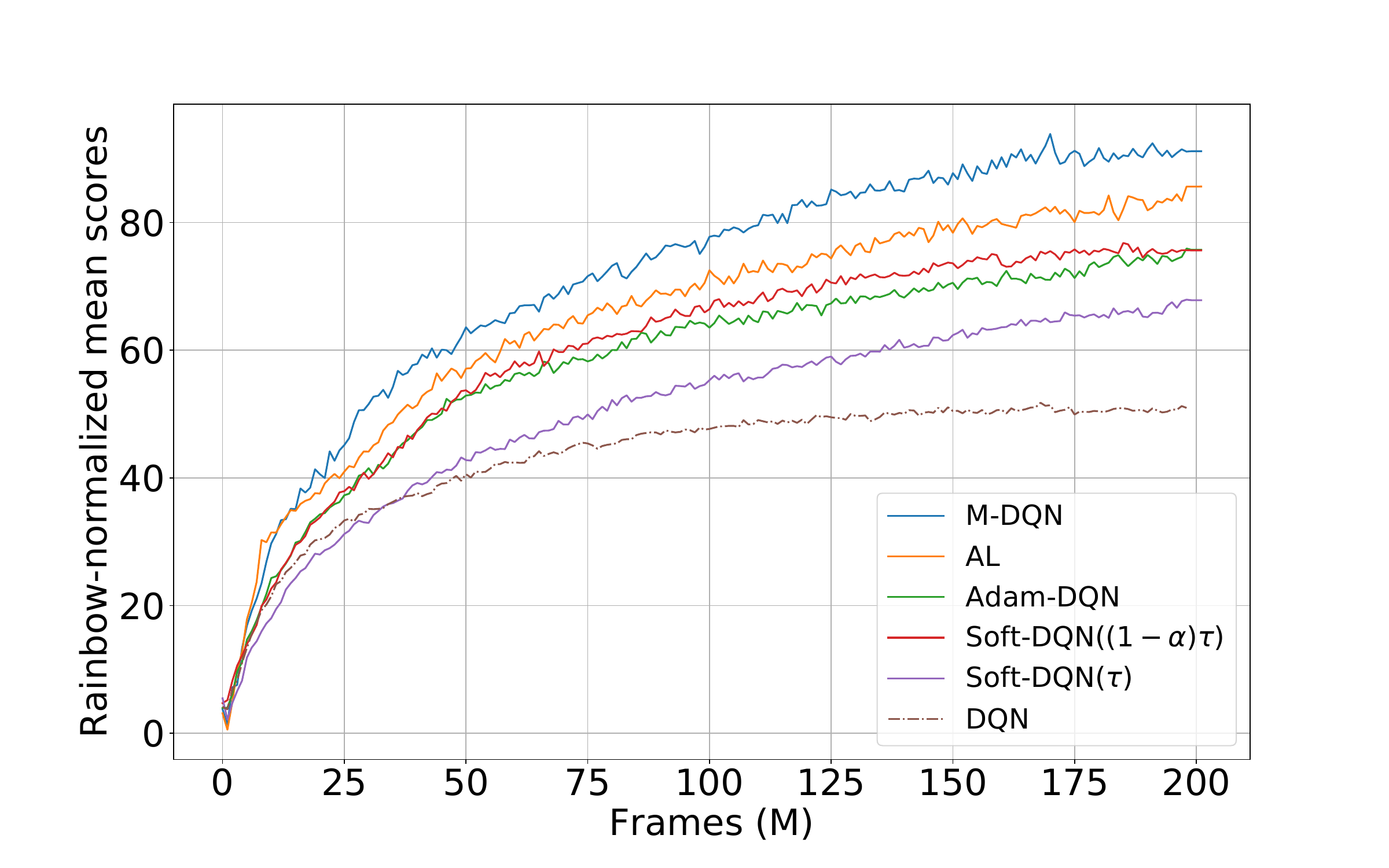}
    \includegraphics[width=.49\linewidth]{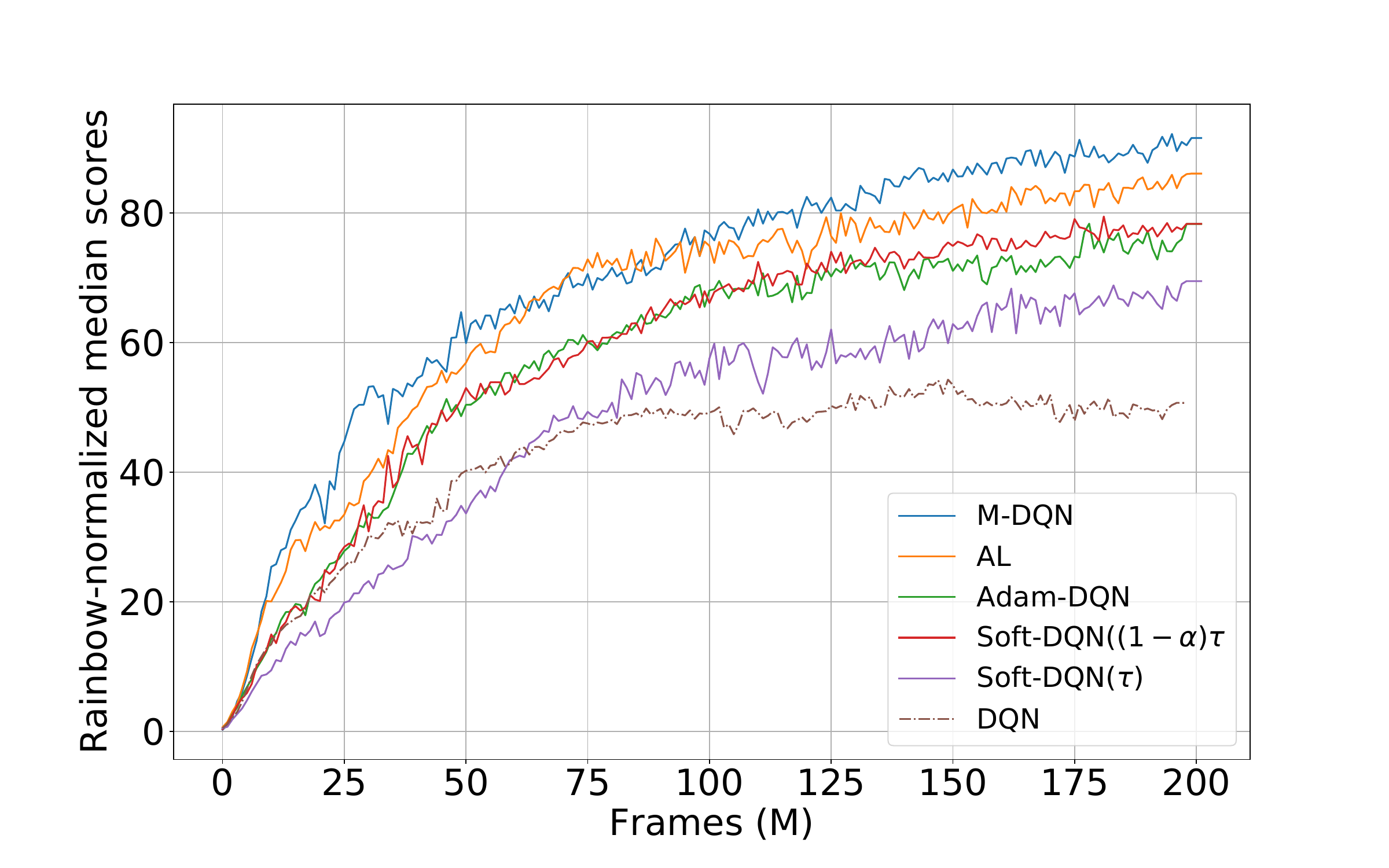}
    \caption{Rainbow-normalized ablation study results.  \textbf{Left:} mean. \textbf{Right:} median.}
    \label{fig:rainbow_ablation}
\end{figure}

We provide complementary results regarding the ablation study:
\begin{itemize}
    \item Fig.~\ref{fig:rainbow_ablation} p.~\pageref{fig:rainbow_ablation} reports the Rainbow-normalized scores of the ablation (instead of the Human-normalized ones in the main paper, Fig.~\ref{fig:ablation}).
    \item Fig.~\ref{fig:auc_multi} p.~\pageref{fig:auc_multi} shows the normalized improvements of all ablations with respect to DQN.
    \item Fig.~\ref{fig:full_ablation} p.~\pageref{fig:full_ablation} reports all learning curves an the 60 Atari games for the ablation.
\end{itemize}

The Rainbow-normalized scores (Fig.~\ref{fig:rainbow_ablation}) confirms the Human-normalized ones (Fig.~\ref{fig:ablation}). The scores themselves are different (due to a different normalization), but the order of the different variations and their gaps is comparable.

Fig.~\ref{fig:auc_multi} provides a summary of the per-game improvement, while Fig.~\ref{fig:full_ablation} provides all related learning curves (Fig.~\ref{fig:auc_multi} summarizing what the results are after 200M frame).
We can observe that M-DQN is not always the best performing agent. Yet, it is very often competitive with the best performing ablation (when M-DQN does not perform the best), and the ablation that surpasses M-DQN is highly game-dependent. Overall, M-DQN is consistently the best performing agent over the whole suite of games, as confirmed by Fig.~\ref{fig:ablation} or Fig.~\ref{fig:rainbow_ablation} both in mean and median Rainbow and Human-normalized scores.

AL performs pretty well (even if less well than M-DQN). Yet, Munchausen-RL is more general, as it consists only in adding a scaled log-policy term to the reward. We've shown in the main paper how it can be readily applied to agents that does not even consider stochastic policies. On the converse, ALE relies heavily on being able to compute the maximum $Q$-value, something which could not be easily extended to continuous actions, contrary to the Munchausen principle. We let this as an interesting direction for future work.
In both average and mean (Fig.~\ref{fig:ablation} and~\ref{fig:rainbow_ablation}), Soft-DQN is the worst ablation, despite being much better in a few games (for example, Amidar or Jamesbond). Again, the temperature was not specifically tuned for Soft-DQN, but it is on par with the close literature (see discussion in Sec.~\ref{sec:experiments}). This suggests that the maximum entropy RL principle alone might not be sufficient, especially when one observes the significant improvement that the Munchausen term brings to it (or, implicitly, adding KL regularization to the entropy term).
We also notice again that Adam DQN works surprisingly well, compared to the original DQN. This is a very interesting finding, and it suggests that Adam DQN should be considered as a better baseline than the seminal DQN.

\subsection{Additional comparison results}
\label{subappx:metrics}

For completeness, we provides additional comparison results:
\begin{itemize}
    \item In addition to the Human-normalized results of Fig.~\ref{fig:human_normalized}, we provide a Rainbow-normalized comparison of the Munchausen agents with respect to DQN, C51, IQN and Rainbow in Fig.~\ref{fig:human_normalized}.
    \item In addition to the per-game normalized improvement of a Munchausen agent with respect to its natural baseline (Fig.~\ref{fig:aucs}), we provide the per-game improvement for M-DQN over DQN, C51, IQN and Rainbow in Fig.~\ref{fig:super_bars_dqn}, as well as the per-game improvement of M-IQN over the same baselines in Fig.~\ref{fig:super_barps_iqn}.
    \item We provide a summary of all best scores (among training, averaged over 3 seeds), for all games on all agents, in Table~\ref{tab:score_all} p.~\pageref{tab:score_all}. M-IQN obtains the most highest-ranking scores among all the considered baselines (including the human one).
    \item For completeness, we report all learning curves of the Munchausen agents and the considered baselines, for the full set of Atari games, in Fig.~\ref{fig:full}.
\end{itemize}
These additional results confirm the observations made in the main paper.

\begin{figure}
    \centering
    \includegraphics[width=0.49\linewidth]{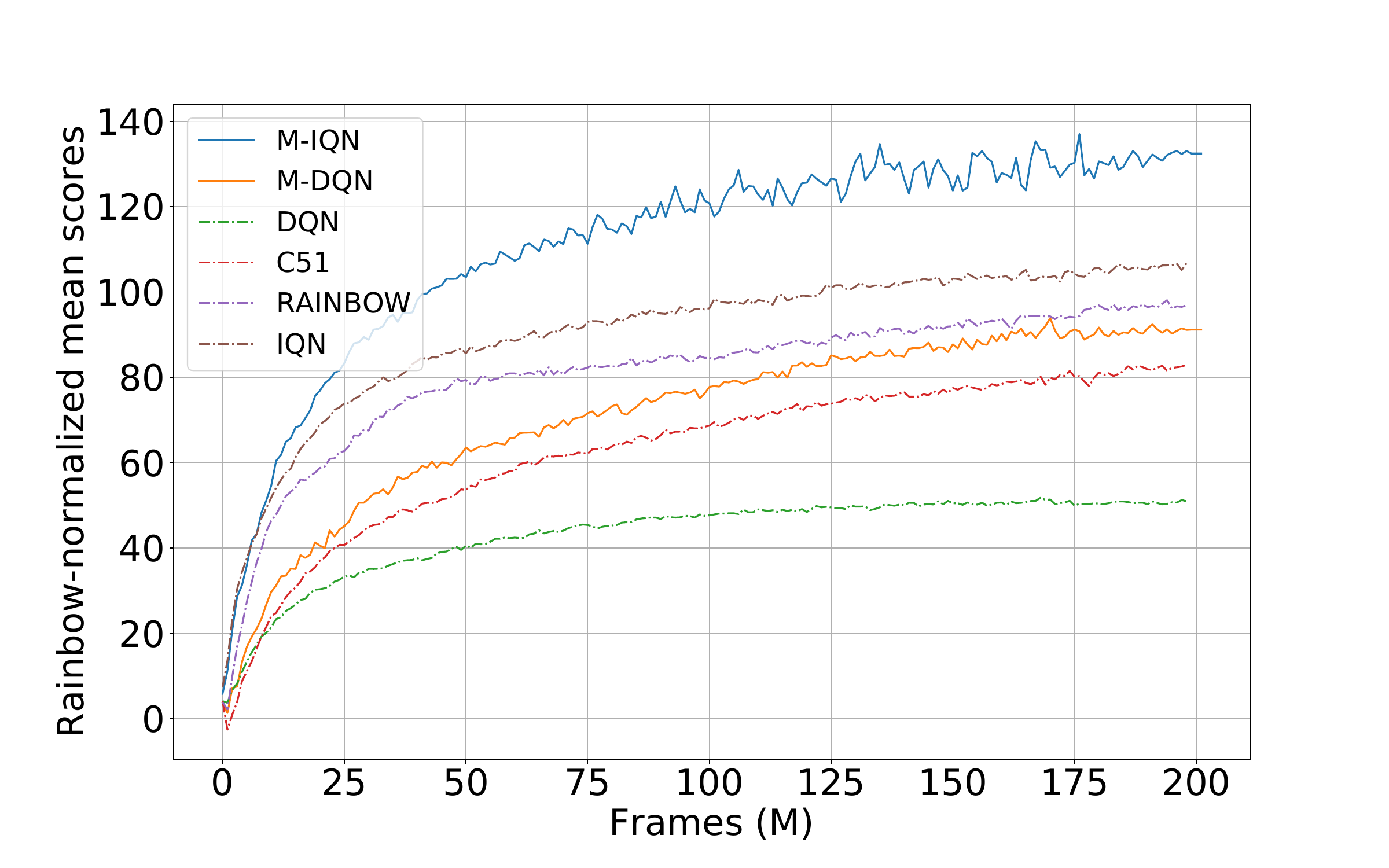}
    \includegraphics[width=0.49\linewidth]{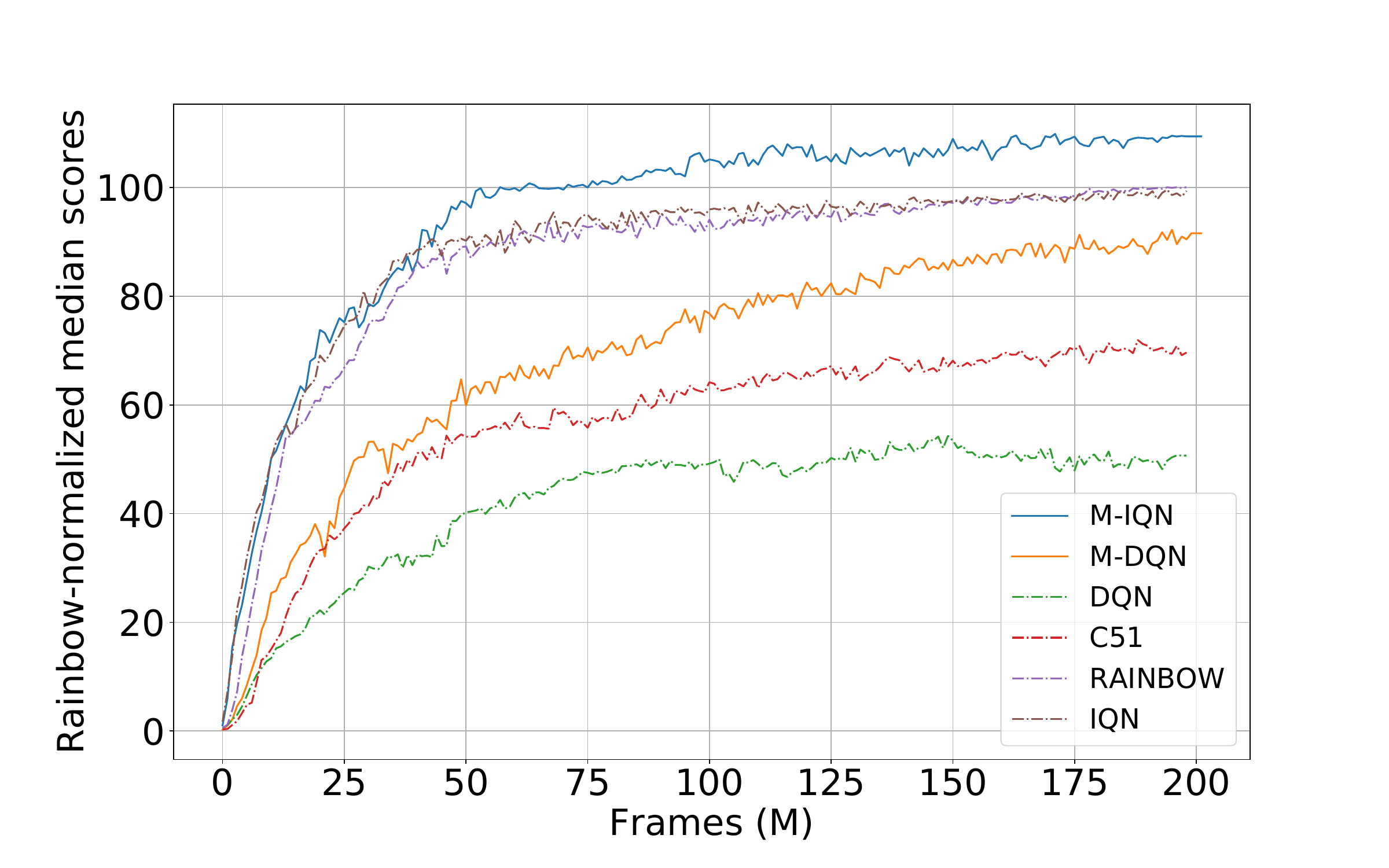}
    \caption{Rainbow-normalized scores. \textbf{Left:} mean. \textbf{Right:} median.}
    \label{fig:rainbow_normalized}
\end{figure}

\begin{figure}
    \centering
    \includegraphics[width=\linewidth]{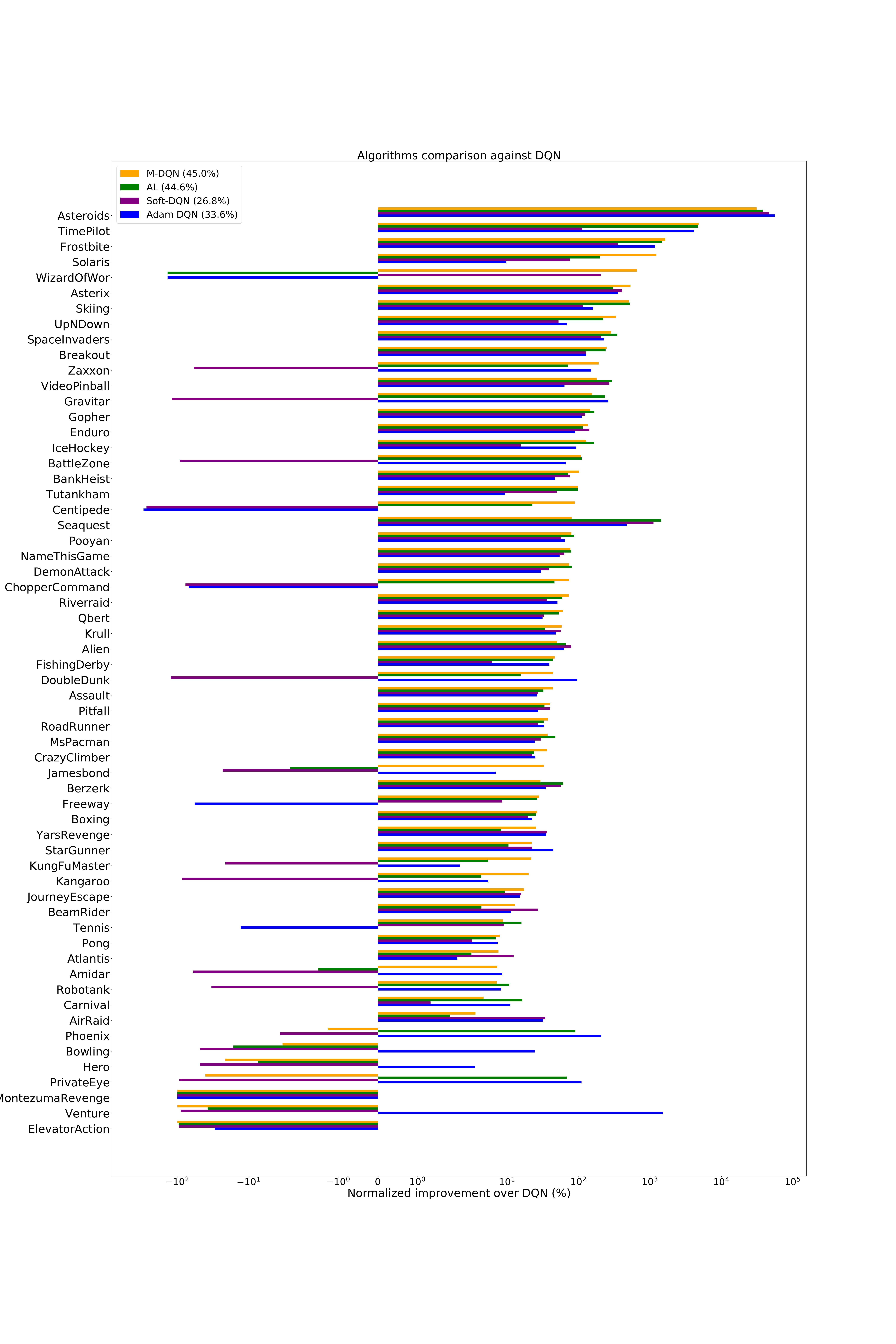}
    \caption{Per games N.I./DQN of the ablation study.}
    \label{fig:auc_multi}
\end{figure}

\begin{figure}
\begin{center}
\includegraphics[width=\textwidth]{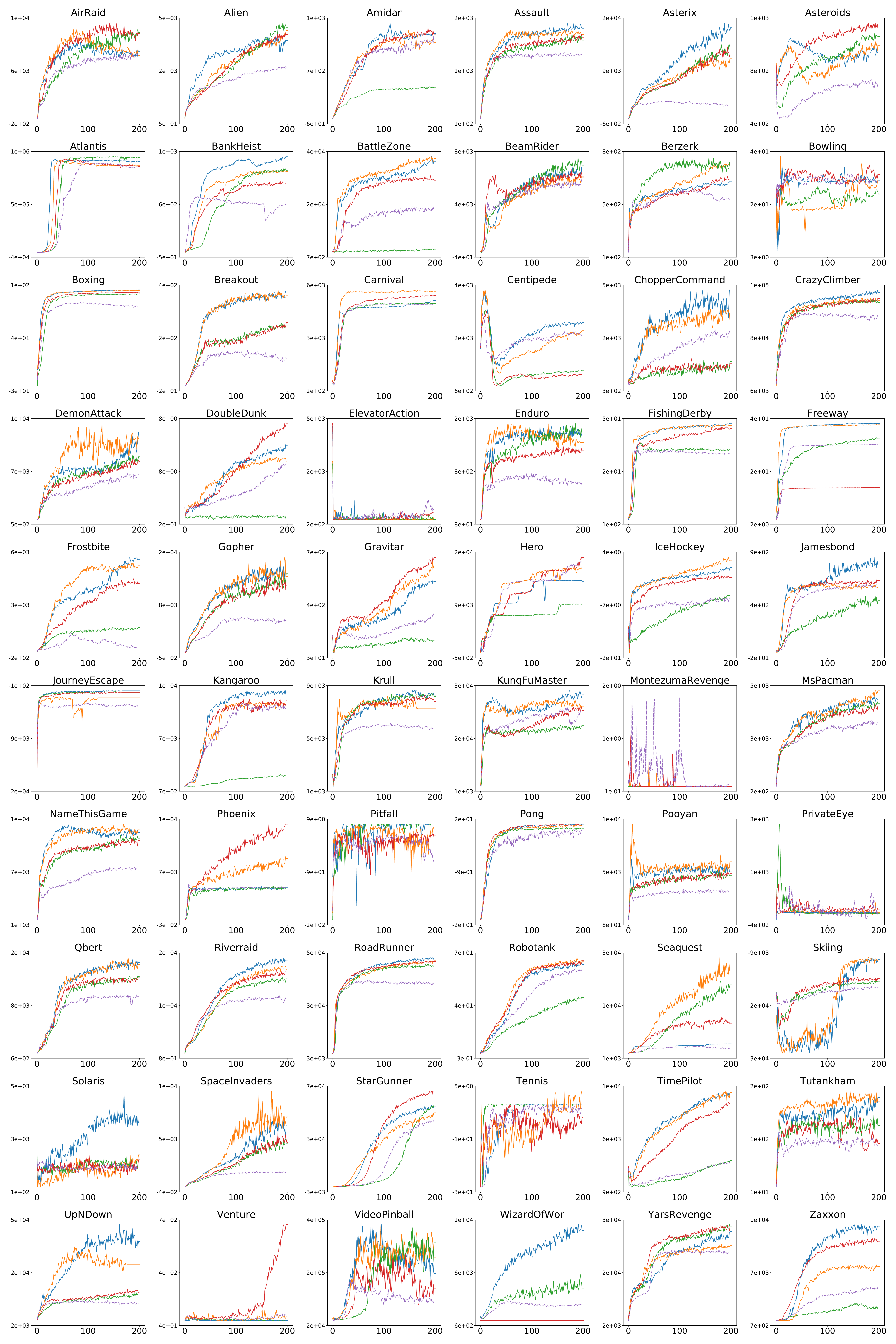}
\caption{All averaged training scores of the ablation. M-DQN in \textcolor{NavyBlue}{blue}, AL in \textcolor{Orange}{orange}, Soft-DQN in \textcolor{ForestGreen}{green}, DQN Adam in  \textcolor{BrickRed}{red}, and DQN in dashed \textcolor{Purple}{purple}.\label{fig:full_ablation}}
\end{center}
\end{figure}

\begin{figure}
    \centering
    \includegraphics[width=\linewidth]{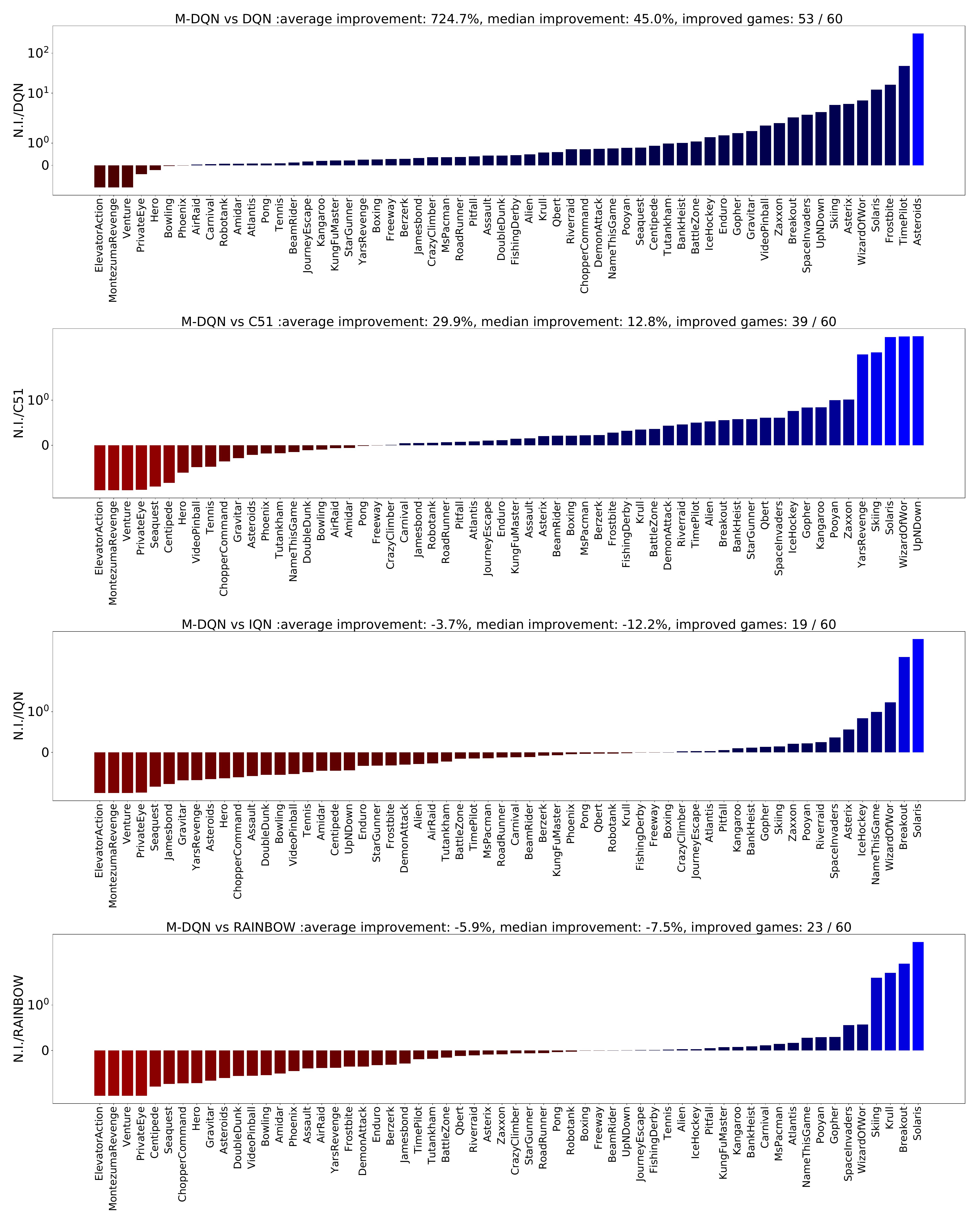}
    \caption{Normalized Improvement of M-DQN vs DQN, C51, IQN, and Rainbow.}
    \label{fig:super_bars_dqn}
\end{figure}

\begin{figure}
    \centering
    \includegraphics[width=\linewidth]{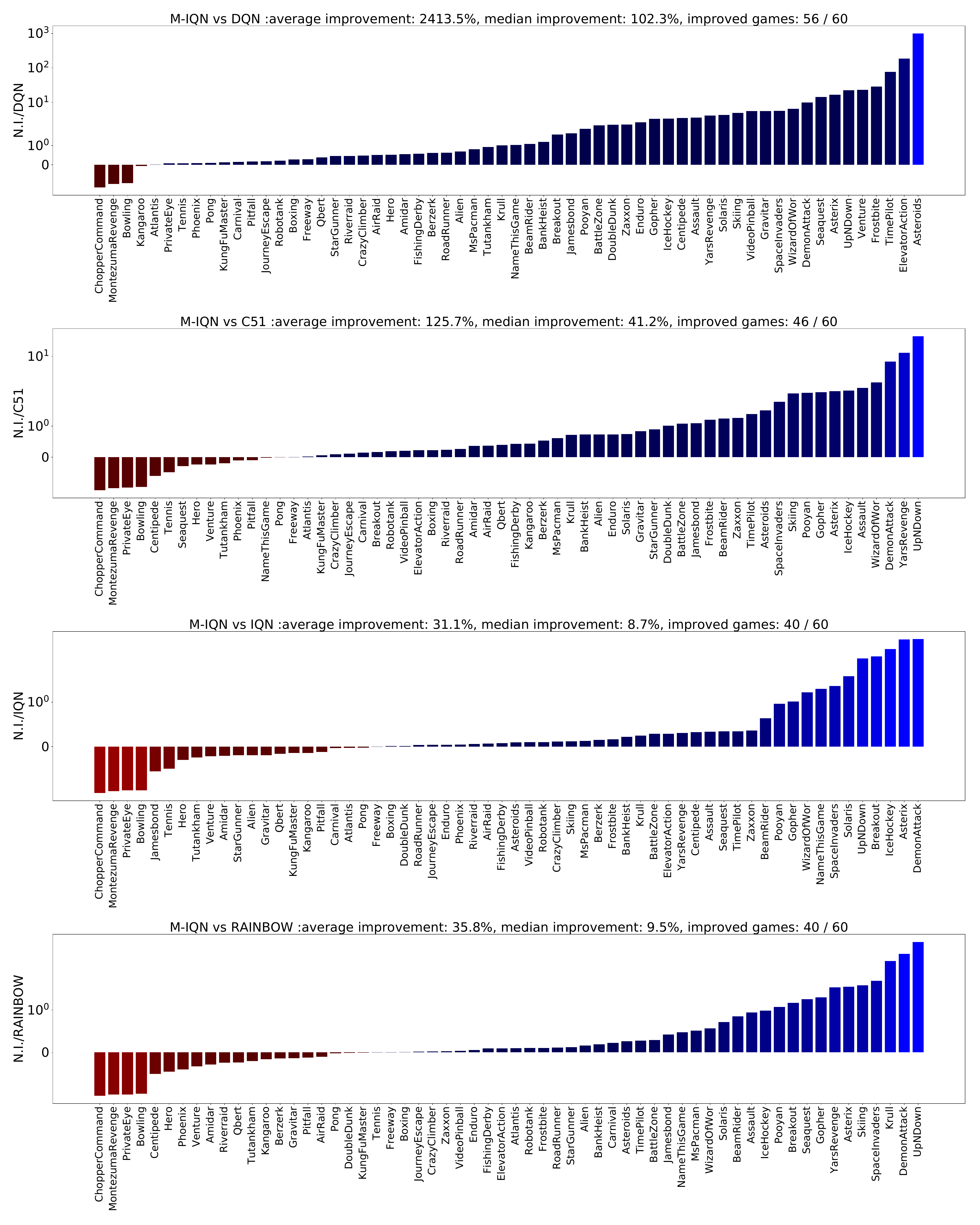}
    \caption{Normalized Improvement of M-IQN vs DQN, C51, IQN, and Rainbow.}
    \label{fig:super_barps_iqn}
\end{figure}

\begin{table}
    \caption{Maximum scores obtained during training (averaged over $100$ episodes and  $3$ random seeds). The bottom line counts the number of games on which an algorithm or a human performs the best.} 
    \centering
\tiny
    \begin{tabular}{l r r r r r r r}
\toprule
&random&human&IQN&DQN&RAINBOW&M-DQN&M-IQN\\ 
\midrule
AirRaid&400&3000&15077&7700&14056&8914&\textbf{19111}\\
Alien&228&\textbf{7128}&5119&2533&3587&3795&4492\\
Amidar&6&1720&2442&1222&\textbf{2630}&1423&1875\\
Assault&222&742&4902&1573&3511&2165&\textbf{7504}\\
Asterix&210&8503&10965&3433&18367&17238&\textbf{49865}\\
Asteroids&719&\textbf{47389}&1616&828&1489&1150&1685\\
Atlantis&12850&29028&893764&919622&838590&\textbf{939533}&918183\\
BankHeist&14&753&1073&704&1148&1190&\textbf{1292}\\
BattleZone&2360&37188&41475&18667&40895&36509&\textbf{52517}\\
BeamRider&364&\textbf{16926}&7365&5852&6529&6745&12775\\
Berzerk&124&\textbf{2630}&662&559&842&608&736\\
Bowling&23&\textbf{161}&46&33&49&37&32\\
Boxing&0&12&98&82&99&98&\textbf{99}\\
Breakout&2&30&159&127&120&\textbf{331}&320\\
Carnival&380&4000&\textbf{5712}&4860&5069&5022&5588\\
Centipede&2091&\textbf{12017}&3816&3337&6618&4134&4371\\
ChopperCommand&811&7388&9301&2852&\textbf{12844}&4507&4573\\
CrazyClimber&10780&35829&137201&109635&147743&140156&\textbf{150783}\\
DemonAttack&152&1971&15433&6411&17802&12114&\textbf{68825}\\
DoubleDunk&-19&-16&21&-6&\textbf{22}&0&22\\
ElevatorAction&0&3000&67224&1723&79968&4215&\textbf{89237}\\
Enduro&0&860&2270&815&2230&1643&\textbf{2332}\\
FishingDerby&-92&-39&45&9&43&44&\textbf{55}\\
Freeway&0&30&34&26&34&34&\textbf{34}\\
Frostbite&65&4335&8061&1186&8572&5453&\textbf{9538}\\
Gopher&258&2412&12108&6044&10641&14728&\textbf{27469}\\
Gravitar&173&\textbf{3351}&1350&330&1272&550&1134\\
Hero&1027&30826&36583&17330&\textbf{46764}&13824&26037\\
IceHockey&-11&1&-0&-6&2&0&\textbf{12}\\
Jamesbond&29&303&\textbf{3596}&589&1106&814&1637\\
JourneyEscape&-18000&-1000&-1252&-2668&-959&-938&\textbf{-806}\\
Kangaroo&52&3035&12872&12192&13460&\textbf{14067}&10939\\
Krull&1598&2666&8910&6410&6229&8912&\textbf{10703}\\
KungFuMaster&258&22736&\textbf{33348}&24495&27900&29607&27119\\
MontezumaRevenge&0&\textbf{4753}&500&2&500&0&0\\
MsPacman&307&\textbf{6952}&5225&3471&4027&4544&6029\\
NameThisGame&2292&8049&9129&7348&9229&11807&\textbf{12761}\\
Phoenix&761&7243&5137&5651&\textbf{8605}&5140&5327\\
Pitfall&-229&\textbf{6464}&-3&-17&-1&0&0\\
Pong&-21&15&\textbf{20}&17&20&19&19\\
Pooyan&500&1000&5339&3535&5640&6396&\textbf{13096}\\
PrivateEye&25&\textbf{69571}&6852&1004&21532&121&100\\
Qbert&164&13455&16995&10399&\textbf{18503}&16415&14739\\
Riverraid&1338&17118&15554&12051&\textbf{21091}&19346&16271\\
RoadRunner&12&7845&59443&39468&55300&51866&\textbf{61269}\\
Robotank&2&12&67&61&66&66&\textbf{73}\\
Seaquest&68&\textbf{42055}&19170&2133&11362&2666&23885\\
Skiing&-17098&\textbf{-4337}&-11035&-15712&-20518&-9671&-10336\\
Solaris&1236&\textbf{12327}&2204&1955&2438&5169&5765\\
SpaceInvaders&148&1669&5452&1850&4420&7504&\textbf{13871}\\
StarGunner&664&10250&\textbf{80362}&45015&57909&55100&65757\\
Tennis&-24&-8&\textbf{23}&-0&0&0&0\\
TimePilot&3568&5229&11887&3768&12283&10590&\textbf{15155}\\
Tutankham&11&168&\textbf{256}&132&245&200&207\\
UpNDown&533&11693&74659&10348&39065&45738&\textbf{216080}\\
Venture&0&1188&1430&52&\textbf{1579}&19&1101\\
VideoPinball&0&17668&485551&177488&513484&368930&\textbf{625118}\\
WizardOfWor&564&4756&6208&2597&8201&12517&\textbf{13644}\\
YarsRevenge&3093&54577&85762&24389&45567&29792&\textbf{111583}\\
Zaxxon&32&9173&11761&4825&15089&13905&\textbf{19080}\\
\midrule
Best&0&14&7&0&8&3&\textbf{28}\\
\bottomrule
    \end{tabular}
    \label{tab:score_all}
\end{table}

\begin{figure}
\begin{center}
\includegraphics[width=\textwidth]{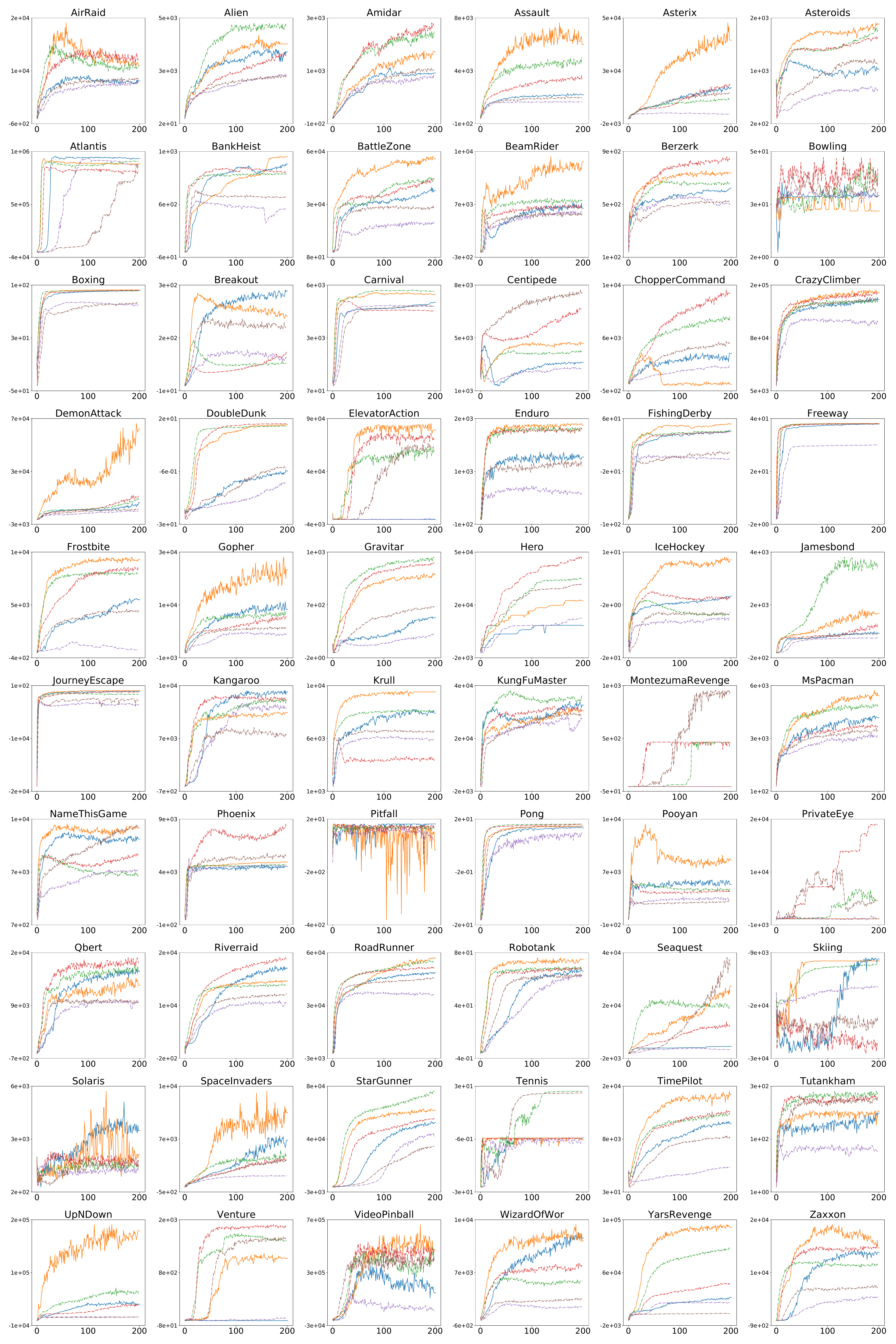}
\caption{All averaged training scores. M-DQN in \textcolor{NavyBlue}{blue}, M-IQN in \textcolor{Orange}{orange}, IQN in dashed \textcolor{ForestGreen}{green}, Rainbow in dashed \textcolor{BrickRed}{red}, DQN in dashed \textcolor{Purple}{purple}, and C51 in dashed \textcolor{Brown}{brown}.\label{fig:full}}
\end{center}
\end{figure}

\end{document}